\documentclass[10pt,twocolumn,letterpaper]{article}

\usepackage[pagenumbers]{iccv} 

\definecolor{iccvblue}{rgb}{0.21,0.49,0.74}
\usepackage[pagebackref,breaklinks,colorlinks,allcolors=iccvblue]{hyperref}

\usepackage{wrapfig}
\usepackage{graphicx}
\usepackage{booktabs}
\usepackage{times}
\usepackage{epsfig}
\usepackage{amsmath,amssymb}
\usepackage{tikz}
\usepackage{comment}
\usepackage{color}
\usepackage{adjustbox}
\usepackage{multirow}
\usepackage{cuted}
\usepackage{mathtools}
\usepackage{kotex}
\usepackage[boxed, ruled,linesnumbered]{algorithm2e}
\usepackage{algpseudocode}
\usepackage{float}
\usepackage{hhline}
\usepackage{boldline}
\usepackage{colortbl}
\usepackage{booktabs} 
\usepackage{bbm}
\usepackage{titletoc}

\usepackage{titling}
\usepackage{caption}

\newcommand{\myparagraph}[1]{\vspace{2pt}\noindent{\bf #1}}

\def\paperID{7730} 
\def\confName{ICCV}
\def\confYear{2025}

\title{\textbf{FaceShield: Defending Facial Image against Deepfake Threats}}

\author{
Jaehwan Jeong$^{1}$ \quad
Sumin In$^{1}$ \quad
Sieun Kim$^{1}$ \quad
Hannie Shin$^{1}$ \quad
Jongheon Jeong$^{1}$ \\
Sang Ho Yoon$^{2}$ \quad
Jaewook Chung$^{3}$ \quad
Sangpil Kim$^{1}$\thanks{Corresponding author} \\ \\
$^{1}$Korea University \quad
$^{2}$KAIST \quad
$^{3}$Samsung Research
}

\date{} 

\begin{document}
\maketitle
\begin{abstract}
The rising use of deepfakes in criminal activities presents a significant issue, inciting widespread controversy. 
While numerous studies have tackled this problem, most primarily focus on deepfake detection.
These reactive solutions are insufficient as a fundamental approach for crimes where authenticity is disregarded.
Existing proactive defenses also have limitations, as they are effective only for deepfake models based on specific Generative Adversarial Networks (GANs), making them less applicable in light of recent advancements in diffusion-based models.
In this paper, we propose a proactive defense method named \textbf{FaceShield}, which introduces novel defense strategies targeting deepfakes generated by Diffusion Models (DMs) and facilitates defenses on various existing GAN-based deepfake models through facial feature extractor manipulations. Our approach consists of three main components: (i) manipulating the attention mechanism of DMs to exclude protected facial features during the denoising process, (ii) targeting prominent facial feature extraction models to enhance the robustness of our adversarial perturbation, and (iii) employing Gaussian blur and low-pass filtering techniques to improve imperceptibility while enhancing robustness against JPEG compression.
Experimental results on the CelebA-HQ and VGGFace2-HQ datasets demonstrate that our method achieves state-of-the-art performance against the latest deepfake models based on DMs, while also exhibiting transferability to GANs and showcasing greater imperceptibility of noise along with enhanced robustness.
\end{abstract}
\vspace{-10pt} 
\section{Introduction}
\label{sec:introduction}
\begin{figure}[ht]
    \vspace{0pt}
    \centering
    \includegraphics[width=0.9\columnwidth]{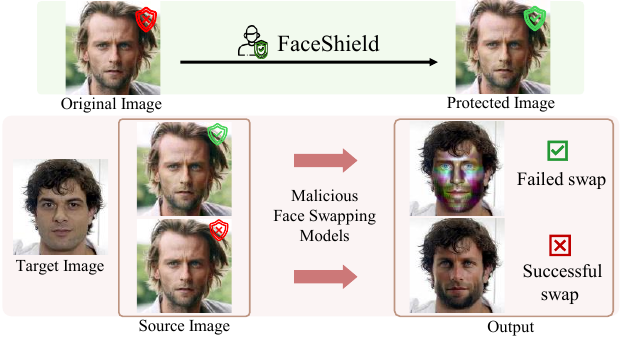}
    \vspace{-2pt}
    \caption{\textbf{Protecting Face during Deepfake using \textit{FaceShield}}. Pure images are vulnerable to face swapping, allowing the target image's face to be easily reflected. In contrast, images protected by \textit{FaceShield} conceal facial feature from deepfake.
    Code is available here: \href{https://github.com/kuai-lab/iccv25_faceshield}{https://github.com/kuai-lab/iccv25\_faceshield}
    }
    \label{fig:showcase}
    \vspace{-15pt}
    
\end{figure}
\begin{figure*}[t]
    \centering
    \includegraphics[width=\textwidth]{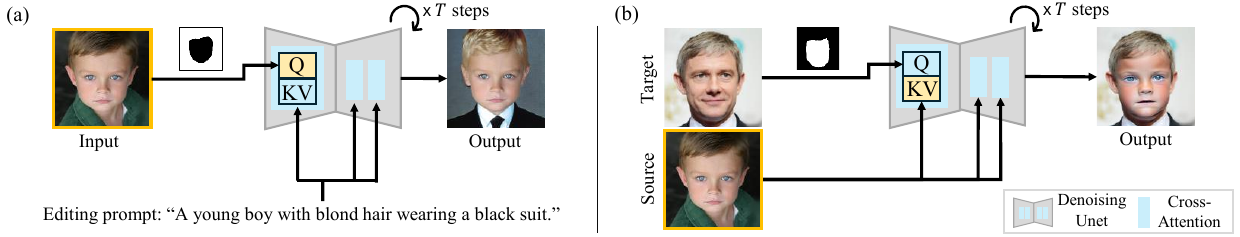}
    \caption{\textbf{Image editing and Deepfake processes in DMs}. (a) In DM-based image editing, a single image is input as a query $Q$ and edited based on a prompt condition. (b) In DM-based deepfake, two images are used, with the target image serving as the query $Q$ while the source image acts as the condition for swapping. This condition operates as key $K$ and value $V$ in the cross-attention layer.}
    \label{fig:motivation}
    \vspace{-10pt}
\end{figure*}

The advancement of deepfake technology and improved accessibility~\cite{li2019faceshifter, Liu2020DeepfacelabIF, zhu2021one, chen2020simswap, wang2021hififace} has led to significant transformations in modern society.
Due to the ease of face swapping, it has been applied across various fields, providing both entertainment and convenience.
However, its powerful capability to generate realistic content has also enabled malicious users to exploit it for criminal purposes, leading to the creation of fake news and various societal problems.

To address the growing concerns surrounding deepfake technology, various countermeasures have been explored, which can be broadly divided into two categories. The first is deepfake detection techniques~\cite{tan2023learning, huang2023implicit, guo2023hierarchical, wang2023noise, choi2024exploiting}, which act as passive defenses by classifying whether content is synthetic or authentic. 
While effective for authenticity verification, these offer only binary results and fail to address advanced threats, such as crimes using realistic fakes.
In contrast, proactive defense strategies offer a more comprehensive solution. These approaches involve embedding imperceptible adversarial perturbation into face images to prevent the protected face from being effectively processed by deepfakes. However, most previous research~\cite{yeh2020disrupting, ruiz2020disrupting, dong2021visually, huang2022cmua, wang2022anti} has concentrated on GAN-based models, often targeting individual models, which limits effectiveness against emerging DM-based deepfakes~\cite{kim2212diffface,zhao2023diffswap,wang2024face,ye2023ip}.
Although significant research exists on image protection within DMs~\cite{liang2023adversarial, shan2023glaze, salman2023raising, liang2023mist, xue2023toward, choi2024diffusionguard} for image editing, the focus has primarily been on attacking the noising and denoising processes when an image is used as a query (Fig.\ref{fig:motivation}a). This leads to targeting the encoder or predicted noise post-UNet processing. However, we observe that such strategies are ineffective for DM-based deepfake models, where the source image influences the output in the form of key-value pairs through attention mechanisms (Fig.\ref{fig:motivation}b).

In this paper, we focus on attacking state-of-the-art DM-based deepfakes while ensuring applicability to GAN-based models as well (Fig.\ref{fig:showcase}).
Given the uncertainty surrounding deepfakes that malicious users might employ, we explore approaches to improve the extensibility across different architectures (e.g., GANs, DMs) and model transferability across different pre-trained backbones. Simultaneously, we propose a novel noise update method that enhances imperceptibility while being robust to JPEG compression.

For DM-based deepfake attacks, we leverage the structural properties in which the conditioning image is embedded and integrated into the denoising UNet through attention mechanisms. By utilizing the IP-Adapter~\cite{ye2023ip}, commonly employed for inpainting, we extract effective adversarial noise from the embedding of the conditioning image. This perturbation effectively disrupts the propagation of the conditioning process, ensuring that the final output does not replicate the features of the protected image.

To enhance the generalizability of our approach, we target two commonly used facial feature extractors. 
First, we attack the MTCNN model~\cite{zhang2016joint}, which uses a cascade pyramid architecture to achieve superior performance and robust detection capabilities. Due to this, it is widely adopted not only in deepfake generation but also across various applications. We leverage the fact that the model scales images to different sizes during face detection. Our perturbation is designed to ensure robustness across various scaling factors and interpolation modes (e.g., $\mathtt{BILINEAR}$, $\mathtt{AREA}$), leading to superior performance compared to existing methods~\cite{kaziakhmedov2019real, zhang2022multi}.
Additionally, we target ArcFace~\cite{deng2019arcface}, a widely adopted pre-trained model for facial feature extraction in deepfake applications. 
By incorporating both methods into our work, we ensure that our approach disrupts a range of deepfakes commonly used for facial landmark detection and feature extraction, thereby improving the overall robustness of our method against various deepfake systems.

In the noise updating process, we refine the perturbation using two techniques: \textit{Noise Blur}, which measures differences between adjacent pixels for imperceptible refinement, and \textit{Low-pass filtering}, retaining low-frequency components, enhancing robustness against JPEG compression.

To summarize, our main contributions are as follows:
\begin{itemize}
\item We introduce a novel attack on deepfakes based on diffusion models. To the best of our knowledge, our proposed method is the first attempt to protect images used as conditions while demonstrating robust performance across various deepfake models by targeting common facial feature extractors.
\item We propose a novel noise update mechanism that integrates Gaussian blur technique with the projected gradient descent method, significantly enhancing imperceptibility. Additionally, we implement low-pass filtering to reduce perturbation loss rates during JPEG compression compared to existing methods.
\item We demonstrate that our deepfake attack method is robust across various deepfake models, outperforming previous diffusion attacks by achieving higher distortion with significantly less noise.
\end{itemize}
\section{Related Work}
\label{sec:related}
\myparagraph{Deepfake techniques.}
With advancements in generative models, deepfake technology has evolved into a specialized field focused on facial synthesis. Previous deepfake models, primarily based on GANs, generally follow a three-stage process: face detection and localization, feature extraction, and face swapping. Among these, studies such as~\cite{zhu2021one, xu2022region, gao2021information, xu2022high, yuan2023reliableswap} employ MTCNN~\cite{zhang2016joint} for face detection and landmark extraction, while the majority of deepfake models, including~\cite{li2019faceshifter, chen2020simswap, zhu2021one, li2021faceinpainter, xu2022region, zhu2024stableswap}, leverage ArcFace~\cite{deng2019arcface} for identity feature extraction.
These steps are similarly employed in DM-based deepfakes that have emerged with the progress of diffusion models. Notable examples, including~\cite{kim2212diffface, zhao2023diffswap, wang2024face}, integrate~\cite{deng2019arcface} to maintain identity consistency. However, recent work has focused on leveraging the capabilities of diffusion models to develop face-swapping methods~\cite{ye2023ip} that achieve high performance without explicitly following previous approaches.

\myparagraph{Enhancing model transferability.}
In the research on adversarial attacks, various attempts have been made to improve transferability. \cite{dong2018boosting, wang2021feature} proposed the model ensemble technique, generating adversarial examples using multiple models to enhance their effectiveness on unseen models. \cite{xie2019improving} introduced a method that selectively utilizes specific layers within a model to improve transferability.
Similarly, \cite{inkawhich2020transferable, huang2019enhancing, zhou2018transferable} investigated techniques that manipulate intermediate layer feature distributions or amplify activation values to prevent adversarial noise from overfitting to a particular model.
Furthermore, \cite{chen2024rethinking, zhang2024on} explored the use of multiple pre-trained backbones within similar model architectures to enhance transferability across different backbone networks.
\begin{figure*}[ht]
\centering
\includegraphics[width=\textwidth]{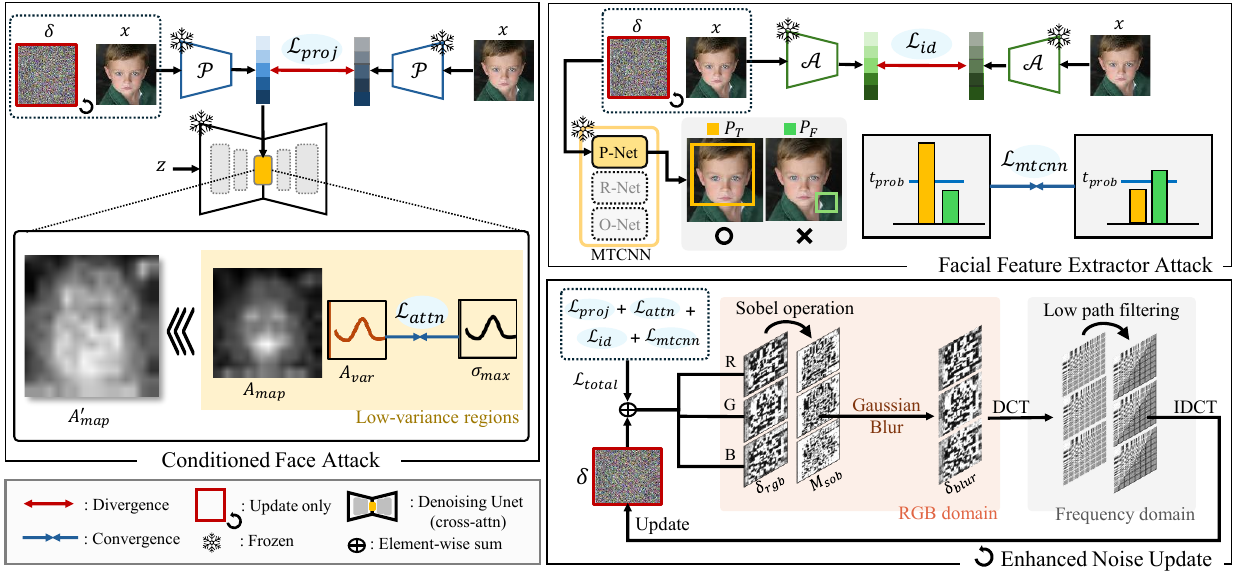}
\vspace{-15pt}
\captionof{figure}
{\textbf{Overview}. Our method has three main parts: (i) Conditioned face attack, which disrupts feature transfer by targeting the embedding process and the attention map variance in the cross-attention layer; (ii) Facial feature extractor attack, which decreases the probability value of face detection and causes extraction disruptions, and (iii) Enhanced noise update, which improves imperceptibility by applying Gaussian blur to regions with significant intensity changes between adjacent pixels, and increases robustness against JPEG compression distortion by encoding the noise in the low-frequency domain.} 
\label{fig:overveiw}
\vspace{-10pt}
\end{figure*}

\section{Method}
\label{sec:methodology}
We propose a novel pipeline, \textit{FaceShield}, to safeguard facial images from being exploited by diverse Deepfake methods through conditional attacks on DMs and facial feature extractor attacks.
In this section, we first introduce the foundational adversarial attack framework utilized across our approach~(Sec.\ref{sec:preliminary}). We then detail our method for disrupting information flow when a facial image is employed as a conditioning input in DMs~(Sec.\ref{sec:diff_attack}). Subsequently, we present our approach for preventing accurate facial feature extraction~(Sec.\ref{sec:fe_attack}). Finally, we introduce our adversarial noise update mechanism, designed to enhance imperceptibility and mitigate degradation from JPEG compression~(Sec.\ref{sec:noise_update}).

\subsection{Preliminaries}
\label{sec:preliminary}

\myparagraph{Cross-attention mechanism.}
To condition generative DMs, the cross-attention mechanism is used, as shown in Fig.\ref{fig:motivation}. Similar to self-attention, it involves computations using the query $Q$, key $K$, and value $V$. However, unlike self-attention, where $Q$, $K$ and $V$ are derived from the same source, cross-attention conditions the process by obtaining $Q$ from the noised image $z_t$ through a learned linear projection $\ell_q$, while $K$ and $V$ are derived from the textual or image embedding $C_\mathtt{emb}$ using learned linear projection $\ell_k$ and $\ell_v$, respectively:
\begin{align}\label{method_eq:2}
    Q = \ell_q(z_t), ~~ K=\ell_k(C_\mathtt{emb}), ~~ V=\ell_v(C_\mathtt{emb}), ~~\text{and} \\
    \mathtt{Attention}(Q, K, V) = \mathtt{Softmax} \left( \frac{Q K^T}{\sqrt{d_k}} \right) V,
\end{align}
where $d_k$ are the dimensions of the key vectors.

\myparagraph{Projected gradient descent (PGD).} 
PGD is a widely used method for crafting adversarial examples when the user has access to the model parameters.
This technique iteratively updates an adversarial perturbation by computing the gradient of a certain loss $\mathcal{L}_\mathtt{adv}$ with respect to the input. At each step, noise is added in the gradient direction while keeping the perturbation within a predefined bound, ensuring the noise is small but effective:
\begin{align}\label{method_eq:3}
    \delta \gets \mathtt{Proj}_{\|\delta\|_\infty \leq \eta} \left( \delta - \alpha \cdot \mathtt{sign}(\nabla_\delta \mathcal{L}_\mathtt{adv}) \right) ,
\end{align}
where $\alpha$ is the step size and $\mathtt{Proj}_{\|\delta\|_\infty \leq \eta}(\cdot)$ projects $\delta$ onto the $\ell_\infty$ ball of radius $\eta$.
By projecting the adversarial example back onto the valid perturbation space, PGD maintains imperceptibility while disrupting the model's predictions.

\subsection{Conditioned Face Attack}
\label{sec:diff_attack}
We now describe our approach to protecting images, specifically by disrupting the effective transfer of information when they are used as conditioning inputs in DMs.
The core of our approach is to effectively interfere with key information using minimal noise, while also ensuring that the model does not overfit by accessing only a minimal number of layers to obtain gradients.
To achieve this, we propose two methods that target both the initial projection phase and the final attention mechanism during the image conditioning process within latent diffusion models~\cite{Rombach_2022_CVPR}.

\myparagraph{Face projector attack.}
When images are used as conditioning inputs, they are firstly transformed into an embedding vector through a pre-trained model~\cite{radford2021learning}. In this method, we access only the topmost layer $\mathcal{P}$ of the model to disrupt the projection process, causing the image to be projected with incorrect information at the initial stage.
For the attack loss function, we consider that converging to a single target value might not ensure consistent convergence speeds or balanced performance. Given that one of our main goals is to design noise applicable to various images, we design our approach to induce random divergence based on the input image, using the $\mathcal{L}_1$ loss function in this process:
\begin{align}\label{method_eq:4}
    \mathcal{L}_{\mathtt{proj}}(\delta ; x) = \| \mathcal{P}(x + \delta) - \mathcal{P}(x) \|_1 ,
\end{align}
where $\delta$ is the adversarial noise.

\myparagraph{Attention disruption attack.}
We also focus on identifying the core vectors within the denoising UNet that are most sensitive to conditional inputs. Initially, we analyze the influence of cross-attention across each UNet layer. Based on prior research~\cite{voynov2023p+}, which shows that different cross-attention layers respond variably to conditioning information, we investigate the impact on perturbation performance for each region.
Our findings lead to the conclusion that targeting attacks near mid-layers produces more significant disruption in qualitative metrics compared to using only the up-down layers or the entire layers, as supported by our experimental results in Fig.\ref{fig:ablation_unet}.
Based on these insights, we propose a novel approach that specifically targets mid-layers during the attack on the diffusion process.

To induce a mismatch in conditioning, we use the mid-layer cross-attention mechanism, as described in Eq. (\ref{method_eq:2}). Based on the idea that the condition is conveyed to the query through attention, we calculate the attention score to obtain the strength of attention.
This is done by performing operations on the query $Q \in \mathbb{R}^{h \times \mathtt{res} \times d}$ and key $K \in \mathbb{R}^{h \times \mathtt{seq} \times d}$, where $h$ (number of heads), $\mathtt{res}$ (resolution), $\mathtt{seq}$ (sequence length), and $d$ (head dimension).
This is followed by a $\mathtt{Softmax}$ operation along the $\mathtt{seq}$ dimension to derive the attention map $A_\mathtt{map} \in \mathbb{R}^{h \times \mathtt{res} \times \mathtt{seq}}$. Exploiting this mechanism, we obtain the variance $A_\mathtt{var}$, allowing us to evaluate attention strength by the following equation:
\vspace{-2pt}
\begin{align}\label{method_eq:5}
    A_\mathtt{var} = \frac{1}{\mathtt{seq}}\sum_{i = 1}^{\mathtt{seq}}\left(A_\mathtt{map}[:, :, i] - \bar{A}_\mathtt{map} \right)^2 \in \mathbb{R}^{h \times \mathtt{res}},
\end{align}
where $\bar{A}_{\mathtt{map}} = \frac{1}{\mathtt{seq}}\sum_{i'=1}^{\mathtt{seq}} A_\mathtt{map}[:, :, i']$ is the mean of attention map across the $\mathtt{seq}$ dimension.

Based on them, we propose an adversarial attack strategy that maximizes $A_\mathtt{var}$, thereby preventing the proper reflection of conditional information $K$ on $Q$.
In this process, we encode the original image $x$ to use as the query $Q$ and project the same $x$ to obtain the key $K$, which is then used to calculate $A_\mathtt{var}$. Thereafter, we find a quantile $P_{t_\mathtt{var}}$ corresponding to a predefined threshold $t_\mathtt{var}$ between 0 and 1 to identify the regions exhibiting weak attention. Using this $P_{t_\mathtt{var}}$, we create a mask $M_\mathtt{var}$ such that values less than or equal to $P_{t_\mathtt{var}}$ are set to 1, and values greater than $P_{t_\mathtt{var}}$ are set to 0. This can be mathematically expressed as follows:
\vspace{-10pt}
\begin{align}
\label{method_eq:6}
    M_\mathtt{var} = \mathbbm{1}[A_\mathtt{var} \le P_{t_\mathtt{var}}],
\end{align}
where $\mathbbm{1}$ is the indicator random variable. 
Subsequently, we derive $A_\mathtt{var}'$ from the same process, using the perturbed image $x + \delta$, and perform attention unequalization on the regions defined by $M_\mathtt{var}$. This method generates missing values by assigning random attention to previously unattended regions between the original images, with the loss function $\mathcal{L}_{\mathtt{attn}}$ defined as follows:
\begin{align}\label{method_eq:7}
    \mathcal{L}_{\mathtt{attn}}( \delta ; \ x, \ \sigma_\mathtt{max})
     = \| (\sigma_\mathtt{max} - A_\mathtt{var}') 
    \odot M_\mathtt{var} \|_2 ,
\end{align}
where $\odot$ is the Hadamard product, and $\sigma_\mathtt{max}$ denotes the maximum variance that can be obtained from the $\mathtt{Softmax}$ output based on the $\mathtt{seq}$, following the equation:
\vspace{-5pt}
\begin{equation}\label{method_eq:8}
    \sigma_\mathtt{max} = \frac{1}{\mathtt{seq}} \left( \left(1 - \frac{1}{\mathtt{seq}}\right)^2 + (\mathtt{seq} - 1) \cdot \frac{1}{\mathtt{seq}^2} \right) .
\end{equation}
In the supplementary material, we provide the algorithm that outlines the method for calculating $\mathcal{L}_{\mathtt{attn}}$.

\subsection{Facial Feature Extractor Attack}
\label{sec:fe_attack}
We design additional perturbation targeting two types of facial extraction, enhancing the applicability of our method not only to DM-based models but also to various other deepfake architecture-based models.

\myparagraph{MTCNN attack.}
We break down the MTCNN attack we propose into three principal stages: (\textit{i}) selecting the resizing scale, (\textit{ii}) enhancing robustness against interpolation, and (\textit{iii}) formulating a loss function to expedite convergence.
Through this process, we achieve not only various resize modes but also model transferability.

Firstly, we select a set of appropriate resizing scales $s_i \in S$. This is to ensure that our attack technique effectively targets only the bounding boxes reaching the final layers of MTCNN. The suitable scale values are selected among the multi-scale factors that the MTCNN model internally uses, and the detailed workings are provided in the supplementary material.

Using the scale factor $s_i$ selected in the previous step, we next scale the input image size $D_\mathtt{adv} = (h,w)$, where $h$ and $w$ are the image's height and width, to yield $D_\mathtt{scl} = (s_i \cdot h, s_i \cdot w)$. This results in an intermediate size $D_\mathtt{int} = D_\mathtt{adv} \odot D_\mathtt{scl}$ obtained through element-wise multiplication. The input image $x \in \mathbb{R}^{c \times h \times w}$ is then upscaled to $D_\mathtt{int}$ using $\mathtt{NEAREST}$ interpolation. Afterward, we downsample the image to $D_\mathtt{scl}$ via average pooling, assigning equal weights to each region referenced during interpolation. This approach runs parallel with a direct $\mathtt{BILINEAR}$ scaling of $D_\mathtt{adv}$ to $D_\mathtt{scl}$, thereby ensuring robust noise generation that functions effectively across various interpolation modes.
 
 In the final stage, we perform a targeted attack on the initial P-Net $\mathcal{T}$ to effectively disrupt the cascade pyramid structure.
 We pass the downsampled adversarial noise-added image $\tilde{x}_\mathtt{adv} \in \mathbb{R}^{c \times s_i \cdot h \times s_i \cdot w}$ through $\mathcal{T}$, which outputs probabilities $P_{\mathtt{T},\mathtt{F}}$ for bounding boxes.
 To expedite the convergence of the MTCNN loss function $\mathcal{L}_\mathtt{mtcnn}$, we propose a masking technique that leverages both the existence probabilities $P_\mathtt{T}$ and the non-existence probabilities $P_\mathtt{F}$. 
 The mask $M_\mathtt{prob}$ is constructed to retain indices in $P_\mathtt{T}$ that exceed the detection threshold $t_\mathtt{prob}$:
\begin{align}\label{method_eq:9}
M_\mathtt{prob} = \mathbbm{1}[P_\mathtt{T}(i,j) > t_\mathtt{prob}],
\end{align}
where $\mathbbm{1}$ is the indicator random variable.
Then, the $\mathcal{L}_\mathtt{mtcnn}$ converges with the mean squared error loss using $M_\mathtt{prob}$: 
\begin{align}\label{method_eq:10}
    \mathcal{L}_{\mathtt{mtcnn}}( \delta ; x, p_\mathtt{gt})
     = \| (\mathcal{T}(x +\delta) - p_\mathtt{gt} ) \odot M_\mathtt{prob} \|_2 ,
\end{align}
where $\mathcal{T}(\cdot) = [P_\mathtt{F}, P_\mathtt{T}]^T$, $p_\mathtt{gt} = [t_\mathtt{prob} + \beta t_\mathtt{prob} - \beta]^T$, and $\beta$ is a value between 0 and 1.
Additional details are provided along with the algorithm in the supplementary material.

\myparagraph{Identity attack.}
To effectively disrupt the accurate reflection of source face information, we target the ArcFace $\mathcal{A}$ models \cite{deng2019arcface}, which are face identity embedding models widely used in deepfake systems. To improve transferability, we ensemble the most commonly used pre-trained backbones within these models.
Since $\mathcal{A}$ represents feature vectors extracted from the same person's face as vectors pointing in similar directions, we designed our approach to induce divergence from the original image $x$ by employing cosine similarity loss, thereby effectively obscuring the relevant identity information:
\begin{align}\label{method_eq:11}
    \mathcal{L}_{\mathtt{id}}(\delta ; x) = \frac{\mathcal{A}(x + \delta) \cdot \mathcal{A}(x)}{\|\mathcal{A}(x + \delta)\|_2 \|\mathcal{A}(x)\|_2} - 1.
\end{align}

\myparagraph{Overall loss operation.} Accordingly, the total loss function $\mathcal{L}_\mathtt{total}$ is defined and used as follows:
\begin{align}\label{method_eq:12}
    \mathcal{L}_\mathtt{total} = \ & \lambda_\mathtt{proj} \mathcal{L}_\mathtt{proj}+ \lambda_\mathtt{attn} \mathcal{L}_\mathtt{attn} \nonumber \\
    & + \lambda_\mathtt{mtcnn} \mathcal{L}_\mathtt{mtcnn} + \lambda_\mathtt{id} \mathcal{L}_\mathtt{id} ,
\end{align}
where each $\lambda$ is a hyperparameter derived from grid searches to control the strength of the respective loss term. Additionally, the sign of $\lambda$ determines the convergence or divergence of the loss function (i.e., $\lambda_\mathtt{proj}$ and $\lambda_\mathtt{id}$ are negative, while $\lambda_\mathtt{attn}$ and $\lambda_\mathtt{mtcnn}$ are positive).

\begin{figure*}[ht]
\centering
\includegraphics[width=\textwidth]{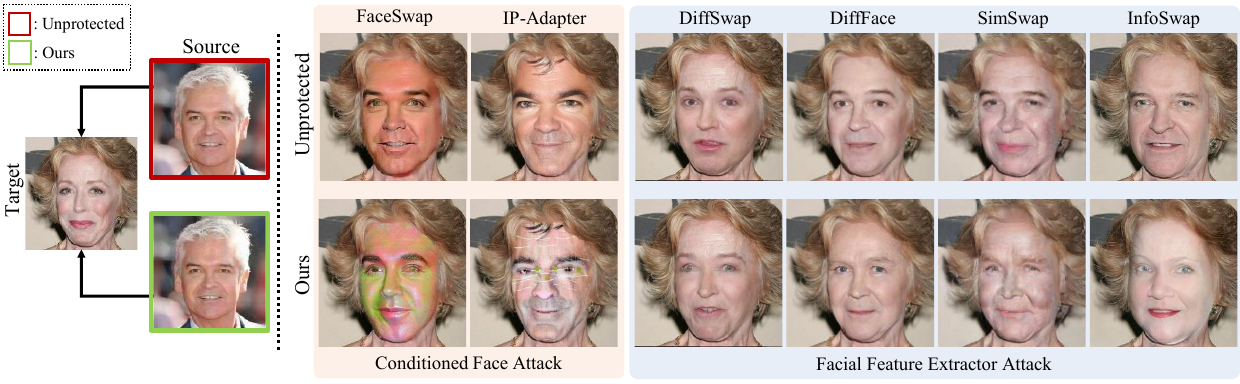}
\vspace{-20pt}
\captionof{figure}
{\textbf{Qualitative Results}. Protection performance across various deepfake models when our adversarial noise is applied. Models~\cite{wang2024face,ye2023ip} highlighted in the orange box typically exhibit facial distortions due to the influence described in Sec.~\ref{sec:diff_attack}, while those~\cite{zhao2023diffswap,kim2212diffface,chen2020simswap,gao2021information} in the blue box display newly generated faces that diverge from the source image, attributed to the impact detailed in Sec.~\ref{sec:fe_attack}.}
\label{fig:ours}
\vspace{-10pt}
\end{figure*}

\subsection{Enhanced Noise Update}
\label{sec:noise_update}
\begin{algorithm}[t!]
\caption{FaceShield}
\label{alg:unified_framework}
\Indm
\KwIn{image $x$, steps $N$, noise clamp $\epsilon$, step size $\alpha$, MTCNN detection threshold $t_\mathtt{prob}$, threshold weight $\beta$, CLIP Image Projector $\mathcal{P}$, Mid-layer cross-attention variance in Stable Diffusion $A_\mathtt{var}'$, MTCNN P-Network $\mathcal{T}$, ArcFace $\mathcal{A}$}
\KwResult{protected image $x_\mathtt{adv}$}
\Indp
Initialize adversarial perturbation $\delta \gets 0$, and protected image $x_\mathtt{adv} \gets x$ \\
\For{n = 1, ..., N}{
    $\mathcal{L}_{\mathtt{proj}} \gets \| \mathcal{P}(x + \delta) - \mathcal{P}(x) \|_1$ \\
    $\mathcal{L}_{\mathtt{attn}} \gets \| (\sigma_\mathtt{max} - A_\mathtt{var}') 
    \odot M_\mathtt{var} \|_2$, \newline
    where $\sigma_\mathtt{max}$ derived from Eq. (\ref{method_eq:8}), and $M_\mathtt{var}$ from Eq. (\ref{method_eq:6}) \\
    $\mathcal{L}_{\mathtt{mtcnn}} \gets \| (\mathcal{T}(x +\delta) - p_\mathtt{gt} ) \odot M_\mathtt{prob} \|_2$, \newline
    where $p_\mathtt{gt} = [t_\mathtt{prob} + \beta t_\mathtt{prob} - \beta]^T$, and $M_\mathtt{prob}$ from Eq. (\ref{method_eq:9}) \\
    $\mathcal{L}_{\mathtt{id}} \gets \frac{\mathcal{A}(x + \delta) \cdot \mathcal{A}(x)}{\|\mathcal{A}(x + \delta)\|_2 \|\mathcal{A}(x)\|_2} - 1$ \\
    Compute the total attack loss: $\mathcal{L}_\mathtt{total} = \lambda_\mathtt{proj} \mathcal{L}_\mathtt{proj}+ \lambda_\mathtt{attn} \mathcal{L}_\mathtt{attn}
     + \lambda_\mathtt{mtcnn} \mathcal{L}_\mathtt{mtcnn} + \lambda_\mathtt{id} \mathcal{L}_\mathtt{id}$ \\
    Update adversarial perturbation: $\delta \gets \alpha \cdot \text{sign}(\nabla_{x_\mathtt{adv}}\mathcal{L}_\mathtt{total})$ \\
    ~~$\delta_\mathtt{blur} \gets \textbf{GaussianBlur}(\delta)$ \\
    ~~$\delta'_\mathtt{rgb} \gets \textbf{LowPassFilter}(\delta_\mathtt{blur})$ \\
    ~~$x_\mathtt{adv} \gets x_\mathtt{adv} - \delta'_\mathtt{rgb}$ \\
    ~~$x_\mathtt{adv} \gets x + \text{clip}(x_\mathtt{adv} - x, -\epsilon, \epsilon)$
}
Clip the image range: $x_\mathtt{adv} \gets \text{clip}(x_\mathtt{adv}, 0, 255)$
\end{algorithm}
\setlength{\textfloatsep}{5pt}
We integrate two additional techniques into the standard PGD to enhance robustness by enabling more imperceptible noise updates and preventing the loss of information due to purification techniques.

\myparagraph{Gaussian blur.}
To enhance noise imperceptibility, we introduce a technique that constrains variations between adjacent regions, addressing the limitations of PGD methods (see Eq. (\ref{method_eq:3})) that only regulate overall noise magnitude. This stems from the observation that differences between neighboring pixels can be as perceptible as the total noise itself. To achieve this, we utilize the Sobel operator~\cite{kanopoulos1988design} to emphasize areas of rapid intensity change, generating a mask $M_\mathtt{sob}$ that highlights image boundaries. Gaussian blur $\mathcal{G}(\cdot)$ is then applied selectively to these regions during noise updates, 
ensuring smoother transitions between adjacent pixels and maintaining a consistent visual appearance:
\begin{align}\label{method_eq:13}
    \delta_\mathtt{blur} = \mathcal{G}(\delta) \odot M_\mathtt{sob} + \delta \odot (1 - M_\mathtt{sob}).
\end{align}

\myparagraph{Low pass filtering.}
To minimize information loss when saving images in $\mathtt{JPEG}$ format and ensure robustness to bit reduction during the compression process~\cite{dziugaite2016study}, low-frequency components are utilized.
At each iteration, the newly updated adversarial noise $\delta_\mathtt{rgb} \in \mathbb{R}^{c \times h \times w}$, where $c$ (channel), $h$ (height), and $w$ (width), undergoes a padding operation and patchification according to a predefined patch size $p$. Then a DCT transform~\cite{1672377} is applied to each patch and channel, resulting in $\delta_\mathtt{dct} \in \mathbb{R}^{c \times h' \times w' \times p \times p}$, where $h' = h / p$ and $w' = w / p$, in the frequency domain.
Using a low-pass filtering mask $M_\mathtt{lp}$, only the low-frequency components of the noise are retained. The noise $\delta'_\mathtt{rgb} \in \mathbb{R}^{c \times h \times w}$ is then reconstructed back into the RGB domain through an inverse transformation. 
The effectiveness of this approach is demonstrated through the experimental results presented in the supplementary material, and the overall operation of 
\textit{FaceShield} is described in Algorithm~\ref{alg:unified_framework}.

\begin{table*}[ht]
\centering
\small
\adjustbox{max width=\textwidth}{
\renewcommand{\arraystretch}{1.1}
\setlength\tabcolsep{2.4pt} 
\begin{tabular}{c| cccc | cccc | cccc | cccc}
    \hlineB{2.5}
    \textbf{Model} & \multicolumn{4}{c}{DiffFace~\cite{kim2212diffface}} & \multicolumn{4}{c}{DiffSwap~\cite{zhao2023diffswap}} & \multicolumn{4}{c}{FaceSwap~\cite{wang2024face}} & \multicolumn{4}{c}{IP-Adapter~\cite{ye2023ip}} \\
    \cline{1-17}
    \textbf{Dataset} & \multicolumn{16}{c}{CelebA-HQ~\cite{karras2017progressive}} \\
    \cline{1-17}
    \textbf{Method} & $L_2$ $\uparrow$ & ISM $\downarrow$ & PSNR $\downarrow$ & HE $\uparrow$ & $L_2$ $\uparrow$ & ISM $\downarrow$ & PSNR $\downarrow$ & HE $\uparrow$ & $L_2$ $\uparrow$ & ISM $\downarrow$ & PSNR $\downarrow$ & HE $\uparrow$ & $L_2$ $\uparrow$ & ISM $\downarrow$ & PSNR $\downarrow$ & HE $\uparrow$ \\
    \hline
    \text{AdvDM~\cite{liang2023adversarial}} & 0.021 & 0.471 & 39.368 & \underline{4.22} & 0.068 & 0.199 & 28.362 & 4.68 & 0.303 & 0.245 & 21.615 & 4.52 & 0.207 & 0.235 & 25.332 & 2.76 \\
    \text{Mist~\cite{liang2023mist}} & 0.021 & 0.468 & 39.443 & 3.94 & 0.067 & 0.201 & 28.384 & 4.18 & 0.287 & 0.230 & 22.263 & \underline{4.78} & 0.152 & 0.265 & 28.213 & 4.26 \\
    \text{PhotoGuard~\cite{salman2023raising}} & 0.022 & 0.469 & 39.194 & 3.82 & 0.068 & 0.201 & 28.292 & 4.58 & 0.282 & 0.238 & 22.316 & 4.44 & 0.153 & 0.268 & 28.101 & \underline{4.44} \\
    \text{SDST~\cite{xue2023toward}} & 0.021 & 0.470 & 39.512 & 4.08 & 0.067 & 0.207 & 28.383 & \underline{5.04} & 0.274 & 0.261 & 22.582 & 4.68 & 0.147 & 0.273 & 28.440 & 4.32 \\
    \hline
    \textbf{Ours} & \textbf{0.044} & \textbf{0.243} & \textbf{32.052} & \textbf{5.76} & \textbf{0.072} & \textbf{0.163} & \textbf{27.833} & \textbf{6.20} & \textbf{0.336} & \textbf{0.194} & \textbf{20.759} & \textbf{6.16} & \textbf{0.350} & \textbf{0.072} & \textbf{20.266} & \textbf{6.60} \\
    \textbf{Ours} (Q=75) & \underline{0.043} & \underline{0.259} & \underline{32.259} & - & \underline{0.070} & \underline{0.169} & \underline{28.034} & - & \underline{0.317} & \underline{0.209} & \underline{21.286} & - & \underline{0.326} & \underline{0.112} & \underline{20.867} & - \\
    \hline
    \hline
    \textbf{Dataset} & \multicolumn{16}{c}{VGGFace2-HQ~\cite{simswapplusplus}} \\
    \cline{1-17}
    \textbf{Method} & $L_2$ $\uparrow$ & ISM $\downarrow$ & PSNR $\downarrow$ & HE $\uparrow$ & $L_2$ $\uparrow$ & ISM $\downarrow$ & PSNR $\downarrow$ & HE $\uparrow$ & $L_2$ $\uparrow$ & ISM $\downarrow$ & PSNR $\downarrow$ & HE $\uparrow$ & $L_2$ $\uparrow$ & ISM $\downarrow$ & PSNR $\downarrow$ & HE $\uparrow$ \\
    \hline
    AdvDM~\cite{liang2023adversarial} & 0.042 & 0.479 & 33.064 & 3.68 & 0.105 & 0.215 & 24.769 & \underline{4.78} & 0.419 & 0.361 & 18.596 & 4.38 & 0.251 & 0.271 & 23.250 & 2.36 \\
    Mist~\cite{liang2023mist} & 0.041 & 0.478 & 33.215 & 4.26 & 0.102 & 0.227 & 24.964 & 3.94 & 0.379 & 0.259 & 19.626 & \underline{4.50} & 0.181 & 0.291 & 26.070 & \underline{4.10} \\
    PhotoGuard~\cite{salman2023raising} & 0.043 & 0.479 & 32.938 & 3.96 & 0.110 & 0.215 & 24.272 & 4.18 & 0.373 & 0.266 & 19.655 & 4.14 & 0.180 & 0.294 & 26.157 & 3.82 \\
    SDST~\cite{xue2023toward} & 0.041 & 0.483 & 33.242 & \underline{5.30} & 0.107 & 0.225 & 24.506 & 4.58 & 0.359 & 0.258 & 19.996 & 4.14 & 0.166 & 0.292 & 26.784 & 4.06 \\
    \hline
    \textbf{Ours} & \textbf{0.062} & \textbf{0.278} & \textbf{29.204} & \textbf{6.10} & \textbf{0.113} & \textbf{0.177} & \textbf{24.054} & \textbf{6.12} & \textbf{0.453} & \textbf{0.237} & \textbf{17.919} & \textbf{6.16} & \textbf{0.382} & \textbf{0.112} & \textbf{19.478} & \textbf{6.42} \\
    \textbf{Ours} (Q=75) & \underline{0.060} & \underline{0.308} & \underline{29.435} & - & \underline{0.112} & \underline{0.185} & \underline{24.201} & - & \underline{0.421} & \underline{0.237} & \underline{18.573} & - & \underline{0.377} & \underline{0.167} & \underline{19.618} & - \\
    \hlineB{2.5}
\end{tabular}
}
\caption{Comparison of perturbation effectiveness among baseline methods on four deepfake models using the CelebA-HQ~\cite{karras2017progressive} and VGGFace2-HQ~\cite{simswapplusplus} datasets.
Our method exhibits the largest distortion in image quality (L2, PSNR) and source similarity (ISM), as well as in human evaluation (HE). Results on JPEG-compressed images (Quality factor 75) further confirm robust protection under compression.}
\vspace{-10pt}
\label{table:comparison}
\end{table*}

\section{Experiments}
\label{sec:experiment}
\subsection{Setups}
\myparagraph{Evaluation details.}
For a fair performance comparison, we use open-source baseline~\cite{liang2023adversarial,liang2023mist,salman2023raising,xue2023toward} and apply noise to the same dataset under identical hyperparameter settings. The corresponding results are presented in Table~\ref{table:noise}, while Table~\ref{table:comparison} provides a performance comparison on diffusion-based deepfakes~\cite{kim2212diffface,zhao2023diffswap,wang2024face,ye2023ip}.
The extensibility experiments on GAN-based models~\cite{chen2020simswap,gao2021information} are shown in Table~\ref{table:gan}, where, in the absence of existing attack methods for these models, we validate \textit{FaceShield}’s effectiveness through comparisons with the original images.
In cases where the feature extractor fails to detect a face, we adjust the generation process to exclude facial features during reconstruction.
Detailed descriptions and an analysis of the resources are provided in the supplementary material.


\myparagraph{Datasets.}
We evaluate our method using two datasets: CelebA-HQ~\cite{karras2017progressive} and VGGFace2-HQ~\cite{simswapplusplus}, both of which have been used in previous studies~\cite{chen2020simswap, wang2024face, gao2021information}. The former is the high-resolution version of CelebA, containing 30,000 celebrity face images, while the latter is the high-resolution version of VGGFace2, consisting of 3.3 million face images from 9,131 unique identities. 
For our experiments, we randomly select 200 identities from each dataset, using 100 images for the source and 100 images for the target.

\subsection{Qualitative Results}
\myparagraph{Performance results across deepfakes.}
As shown in Fig.\ref{fig:ours}, \textit{FaceShield} demonstrates robustness across various deepfake models. The perturbations result in either (i) pronounced artifacts reflecting non-relevant facial information instead of key features~\cite{ye2023ip,wang2024face}, or (ii) a complete misinterpretation of the source face, generating a new, unrelated identity~\cite{kim2212diffface,zhao2023diffswap,chen2020simswap,gao2021information}.

\begin{figure}[ht]
    \centering
    \includegraphics[width=\columnwidth]{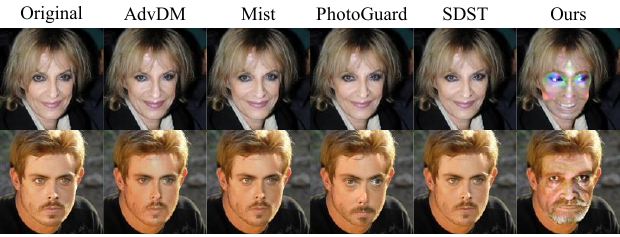}
    \vspace{-10pt}
    \caption{We generate deepfake~\cite{ye2023ip} results from protected images of methods~\cite{liang2023adversarial,liang2023mist,salman2023raising,xue2023toward}. While these fail to disrupt deepfake generation, our method causes deepfakes to malfunction.}
    \label{fig:ip_adapter}
\end{figure}

\myparagraph{Comparison with state-of-the-art methods.}
We compare our method with baselines on DM-based deepfake model~\cite{ye2023ip}. 
Although the methods~\cite{liang2023adversarial,liang2023mist,salman2023raising,xue2023toward} that achieved high performance in diffusion adversarial attacks fail to induce visible changes on the deepfake model, ours demonstrates strong protective performance (Fig.\ref{fig:ip_adapter}).

\subsection{Quantitative Results}
\label{sec:experiment_3}
\myparagraph{Automatic metrics.}
As shown in Table~\ref{table:comparison}, we compare \textit{FaceShield} to baseline methods across deepfake models~\cite{kim2212diffface,zhao2023diffswap,wang2024face,ye2023ip,chen2020simswap,gao2021information} using $L_2$, Identity Score Matching~\cite{van2023anti} (ISM), and PSNR.
The $L_2$ and PSNR metrics evaluate image quality by comparing deepfake results from clean and protected images, with higher $L_2$ and lower PSNR indicating more distortion.
ISM measures the similarity between the source face and the deepfake output, with lower values indicating less similarity.
We conduct experiments on 100 source-target pairs from CelebA-HQ~\cite{karras2017progressive} and VGGFace2-HQ~\cite{simswapplusplus}, showing that \textit{FaceShield} outperforms baselines across all metrics. 
We also analyze the noise levels in protected images using LPIPS, PSNR, and SSIM, as shown in Table~\ref{table:noise}. 
These image quality metrics, compared between protected and original images, show that our method consistently produces less noise than baseline methods.
Additionally, we measure the Frequency Rate (FR), which indicates that most of \textit{FaceShield}'s noise is concentrated in low frequencies. This property helps maintain its effectiveness under JPEG compression. To verify, we compressed the protected images to JPEG Quality 75 and tested across four deepfake models.
The results show that while performance slightly decreases, \textit{FaceShield} still outperforms baseline methods, as shown in Table~\ref{table:comparison}, \textbf{Ours} (Q=75).

\begin{table}[ht]
\centering
\adjustbox{max width=\columnwidth}{
\begin{tabular}{c | ccccc}
    \hlineB{2.5}
    \textbf{Dataset} & \multicolumn{5}{c}{CelebA-HQ~\cite{karras2017progressive}} \\
    \hline
    \textbf{Method} & LPIPS $\downarrow$ & PSNR $\uparrow$ & SSIM $\uparrow$ & FR $\uparrow$ & HE $\uparrow$ \\ 
    \hline
    AdvDM~\cite{liang2023adversarial} & \underline{0.4214} & 30.4476 & 0.8438 & \underline{2.1077} & 3.86 \\
    Mist~\cite{liang2023mist} & 0.5492 & 29.9935 & 0.8684 & 1.6583 & 4.70 \\
    PhotoGuard~\cite{salman2023raising} & 0.5515 & 29.9127 & 0.8669 & 1.6538 & 4.82 \\ 
    SDST~\cite{xue2023toward} & 0.5409 & \underline{31.4762} & \underline{0.9033} & 1.6767 & \underline{5.12} \\
    \rowcolor{gray!30}
    \textbf{Ours} & \textbf{0.2017} & \textbf{32.6289} & \textbf{0.9394} & \textbf{18.4651} & \textbf{5.64} \\
    \hline
    \hline
    \textbf{Dataset} & \multicolumn{5}{c}{VGGFace2-HQ~\cite{simswapplusplus}} \\
    \hline
    \textbf{Method} & LPIPS $\downarrow$ & PSNR $\uparrow$ & SSIM $\uparrow$ & FR $\uparrow$ & HE $\uparrow$ \\ 
    \hline
    AdvDM~\cite{liang2023adversarial} & \underline{0.4108} & 30.2523 & 0.8436 & \underline{2.0667} & 3.66 \\
    Mist~\cite{liang2023mist} & 0.5208 & 29.9068 & 0.8721 & 1.6872 & 4.34 \\
    PhotoGuard~\cite{salman2023raising} & 0.5221 & 29.8204 & 0.8712 & 1.6824 & \underline{4.62} \\ 
    SDST~\cite{xue2023toward} & 0.5060 & \underline{31.3545} & \underline{0.9092} & 1.6892 & 4.48 \\
    \rowcolor{gray!30}
    \textbf{Ours} & \textbf{0.1941} & \textbf{31.5799} & \textbf{0.9341} & \textbf{18.0400} & \textbf{5.28} \\
    \hlineB{2.5}
\end{tabular}}
\caption{With the same step size, iterations, and noise clamping values applied, our method shows the least distortion across three image quality metrics (LPIPS, SSIM, PSNR, HE) and demonstrates a higher low-frequency content (FR).}
\label{table:noise}
\end{table}

\myparagraph{Human evaluation.}
We conduct a human evaluation (HE) on the same methods and models, using 20 source images and 100 participants recruited via Amazon Mechanical Turk. Participants assess two factors: protection noise visibility (Table \ref{table:noise}) and the similarity between the source image and its deepfake output (Table \ref{table:comparison}). 
We employ a Likert scale from 1 to 7. For noise visibility, a score of 7 indicates the least visible noise, while for deepfake similarity, a score of 7 reflects a significant deviation from the source identity. 

\subsection{Ablation Study}
\myparagraph{Effect of gaussian blur on noise.}
To evaluate the effect of Gaussian blur, one of the enhanced noise update methods, we present qualitative results in the Supplementary Material comparing the blur effect's presence and absence. From this comparison, it is clear that the noise update becomes significantly more invisible at the boundaries of abrupt changes in the noise, as detected through Sobel filtering.

\myparagraph{Impact of each loss function.}
To demonstrate the generalizability of each loss function, we conducted an ablation study across six models using the ISM metric, as shown in Table~\ref{table:loss}. The results illustrate how \textit{FaceShield} protects faces, as seen in Table~\ref{table:comparison} and Table~\ref{table:gan}, with \colorbox{red!20}{red} shading indicating performance degradation when a loss function is removed.
This experiment shows that each loss function successfully impacts multiple models, and by combining them into $\mathcal{L}_\mathtt{total}$, we cover a broader range of deepfakes.

\begin{table}[ht]
\vspace{-5pt}
\centering
\small
\adjustbox{max width=\columnwidth}{
    \renewcommand{\arraystretch}{1.1}
    \setlength\tabcolsep{3pt} 
    \begin{tabular}{l|c|c|c|c|c|c}
        \hlineB{2.5}
        \textbf{~~~~~ISM$\downarrow$} & DiffFace & DiffSwap & FaceSwap & IP-Adapter & SimSwap & InfoSwap \\
        \hline
        \text{w/o} $\mathcal{L}_\mathtt{proj}$ & 0.241 & \cellcolor{red!20}{0.167} & \cellcolor{red!20}{0.270} & \cellcolor{red!20}{0.135} & \cellcolor{red!20}{0.544} & \cellcolor{red!20}{0.256} \\
        \text{w/o} $\mathcal{L}_\mathtt{attn}$ & \cellcolor{red!20}{0.254} & \cellcolor{red!20}{0.170} & \cellcolor{red!20}{0.223} & \cellcolor{red!20}{0.076} & 0.168 & \cellcolor{red!20}{0.252}\\
        \text{w/o} $\mathcal{L}_\mathtt{mtcnn}$ & 0.231 & \cellcolor{red!20}{0.174} & 0.166 & 0.047 & 0.183 & \cellcolor{red!20}{0.354} \\
        \text{w/o} $\mathcal{L}_\mathtt{id}$ & \cellcolor{red!20}{0.446} & \cellcolor{red!20}{0.175} & \cellcolor{red!20}{0.217} & 0.040 & \cellcolor{red!20}{0.512} & \cellcolor{red!20}{0.430} \\
        \hline
        ~~~~$\mathcal{L}_\mathtt{total}$ & 0.243 & 0.163 & 0.194 & 0.072 & 0.184 & 0.237 \\
        \hlineB{2.5}
    \end{tabular}
    }
\vspace{-5pt}
\caption{Each model's performance is measured using the ISM score, confirming that each loss function ensures transferability across various deepfakes. As a result, the integrated $\mathcal{L}_\mathtt{total}$ is capable of covering a wider range.}
\label{table:loss}
\end{table}
\vspace{-7pt}

\begin{figure}[h]
    \centering
    \vspace{-7pt}
    \includegraphics[width=\columnwidth]{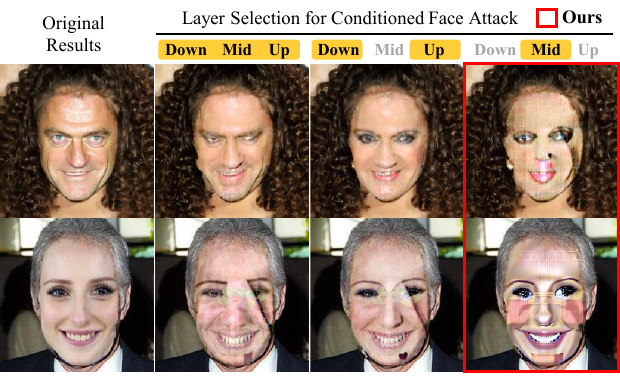}
    \vspace{-20pt}
    \caption{A comparison of conditioned face attack results across UNet~\cite{Rombach_2022_CVPR} layers shows the best protection when targeting mid-layer cross-attention.}
    \label{fig:ablation_unet}
    \vspace{-6pt}
\end{figure}

\myparagraph{Attack effectiveness of mid-layers.}
We qualitatively show that focusing the diffusion attack on the mid-layers of the denoising UNet~\cite{Rombach_2022_CVPR} is more effective than applying it to the entire layers or the up/down layers, as shown in Fig.\ref{fig:ablation_unet}. 
The experiment is conducted by applying noise $\delta=4/255$ to create protected images, which are then passed through the deepfake model~\cite{ye2023ip} to compare the resulting outputs.

\myparagraph{MTCNN resize evaluation.}
To demonstrate the superiority of our method across various resizing modes and model transferability, we conduct an ablation study comparing it to the conventional $\mathtt{BILINEAR}$ scaling method~\cite{zhang2022multi}. We evaluate performance using different scaling methods from the $\mathtt{OpenCV}$ and $\mathtt{PIL}$ libraries, with 3,000 images from both the CelebA-HQ and VGGFace2-HQ datasets. Transferability is assessed through experiments conducted on both PyTorch and TensorFlow versions. The evaluation is based on the final bounding box detection probabilities from MTCNN~\cite{zhang2016joint}, and the results in the supplementary material confirm that our method outperforms existing approaches.

\subsection{Applications}
\myparagraph{Extensibility on GAN-based deepfake models.}
We also conduct additional experiments on the GAN-based diffusion model~\cite{chen2020simswap,gao2021information}. 
The experimental conditions are the same as those in Table~\ref{table:noise}, and the results, as shown in Table~\ref{table:gan}, indicate a degradation in model performance. 
Qualitative assessments are provided in Fig.~\ref{fig:ours}.

\begin{table}[H]
\vspace{-5pt}
\centering
\small
\adjustbox{max width=\columnwidth}{
\begin{tabular}{c | ccc | ccc}
    \hlineB{2.5}
    \textbf{Model} & \multicolumn{3}{c}{SimSwap~\cite{chen2020simswap}} & \multicolumn{3}{c}{InfoSwap~\cite{gao2021information}} \\ 
    \hline
    \textbf{Dataset} & \multicolumn{6}{c}{CelebA-HQ~\cite{karras2017progressive}} \\
    \hline
    \textbf{Method} & $L_2$ $\uparrow$ & ISM $\downarrow$ & PSNR $\downarrow$ & $L_2$ $\uparrow$ & ISM $\downarrow$ & PSNR $\downarrow$ \\
    \hline
    Original & 0.000 & 0.544 & 80.000 & 0.000 & 0.431 & 80.000 \\
    \cellcolor{gray!30}\textbf{Ours} & \cellcolor{gray!30}0.070 & \cellcolor{gray!30}0.184 & \cellcolor{gray!30}26.921 & \cellcolor{gray!30}0.129 & \cellcolor{gray!30}0.237 & \cellcolor{gray!30}30.220 \\
    \hline
    \hline
    \textbf{Dataset} & \multicolumn{6}{c}{VGGFace2-HQ~\cite{simswapplusplus}} \\
    \hline
    \textbf{Method} & $L_2$ $\uparrow$ & ISM $\downarrow$ & PSNR $\downarrow$ & $L_2$ $\uparrow$ & ISM $\downarrow$ & PSNR $\downarrow$ \\
    \hline
    Original & 0.000 & 0.681 & 80.000 & 0.000 & 0.565 & 80.000 \\
    \cellcolor{gray!30}\textbf{Ours} & \cellcolor{gray!30}0.067 & \cellcolor{gray!30}0.314 & \cellcolor{gray!30}27.305 & \cellcolor{gray!30}0.142 & \cellcolor{gray!30}0.356 & \cellcolor{gray!30}29.044 \\
    \hlineB{2.5}
\end{tabular}}
\vspace{-5pt}
\caption{Applicability of \textit{FaceShield} to other deepfake frameworks. Our method, when applied to GAN-based models, not only reduces image quality ($L_2$, PSNR) but also significantly lowers source similarity (ISM).}
\vspace{-10pt}
\label{table:gan}
\end{table}

\myparagraph{Transferability with different weights.}
To demonstrate that \textit{FaceShield} ensures robust transferability not only across structurally different models but also to models with similar architectures but different pre-trained weights, we evaluated its performance on various versions of IP-Adapter~\cite{ye2023ip}. The results, which can be found in the supplementary material, confirm the superior transferability performance of our method.
\section{Conclusion}
\label{sec:conclusion}
In this study, we propose \textit{FaceShield}, an invisible facial protection technique designed to attack various deepfakes. Through comparisons with multiple baseline methods, we demonstrate that \textit{FaceShield} offers superior protection with significantly lower resource costs, particularly for deepfake models utilizing the latest diffusion techniques. Furthermore, its design integrates diverse transferability enhancement strategies, ensuring consistent performance not only across various pretrained versions but also across diffusion-based models with different architectures. This robustness extends to entirely different architectures, including GAN-based models. Additionally, by incorporating an improved noise update mechanism that ensures invisibility while minimizing information loss, \textit{FaceShield} proves to be a practical and effective solution for preventing the misuse of facial images across a wide range of deepfake systems.

\myparagraph{Limitations and Future Work.} Although we introduce a method to enhance robustness against JPEG compression and resizing, other purification techniques still exist, which may lead to the potential loss of our protective noise information. Therefore, we plan to further strengthen the noise to effectively counter a broader range of purification methods.
\clearpage
\section{Acknowledgment}
\label{sec:acknowledgment}

This work was supported by Culture, Sports and Tourism R\&D Program through the Korea Creative Content Agency grant funded by the Ministry of Culture, Sports and Tourism 
(International Collaborative Research and Global Talent Development for the Development of Copyright Management and Protection Technologies for Generative AI, RS-2024-00345025, 47\%;
Research on neural watermark technology for copyright protection of generative AI 3D content, RS-2024-00348469, 25\%), 
the National Research Foundation of Korea(NRF) grant funded by the Korea government(MSIT)(RS-2025-00521602, 25\%),
Institute of Information \& communications Technology Planning \& Evaluation (IITP) \& ITRC(Information Technology Research Center) grant funded by the Korea government(MSIT) (No.RS-2019-II190079, Artificial Intelligence Graduate School Program(Korea University), 1\%; 
IITP-2025-RS-2024-00436857, 1\%; IITP-2025-RS-2025-02304828, Artificial Intelligence Star Fellowship Support Program to Nurture the Best Talents, 1\%), 
and
Artificial intelligence industrial convergence cluster development project funded by the Ministry of Science and ICT(MSIT, Korea)\&Gwangju Metropolitan City.

{
    \small
    \bibliographystyle{ieeenat_fullname}
    \bibliography{main}
}
\clearpage
\setcounter{page}{1}
\setcounter{section}{0}
\renewcommand{\thesection}{\Alph{section}}
\maketitlesupplementary
\vspace{5\baselineskip}
\section*{Contents}
\startcontents[sections]
\printcontents[sections]{c}{1}{\setcounter{tocdepth}{2}}



\clearpage
\section{Additional Explanation on Our Attack}
\subsection{Attention disruption Attack}
\label{sup:a1}
\myparagraph{Algorithm.}
The full procedure of attention disruption attack is summarized in Algorithm~\ref{alg:attn}.

\begin{algorithm}[ht]
    \setlength{\belowcaptionskip}{-20pt}
    \caption{Adversarial loss in cross attention.}
    \label{alg:attn}
    \Indm
    \KwIn{
        perturbation $\delta$, query embedding $Q_x$, original source face embedding $K_x$, adversarial source face embedding $K_{(x+\delta)}$, low variance threshold $t_\mathtt{var}$, maximum variance value $\sigma_\mathtt{max}$, low variance mask $M_\mathtt{var}$, attention loss $\mathcal{L}_\mathtt{attn}$, attention loss function $\mathcal{F}$
    }
    \KwResult{
        stored low-variance mask $M_\mathtt{var}$, added attention loss $\mathcal{L}_\mathtt{attn}$
    }
    \Indp
    \If{$M_\mathtt{var}$ is not precomputed}{
        \tcp{\textcolor{blue}{Construct Ground Truth}}
        Compute original attention map: $A_\mathtt{map} \gets \textbf{Softmax}(Q_x K_x^T / \sqrt{d})$ \\
        Compute variance: $A_\mathtt{var} \gets \textbf{Var}(A_\mathtt{map})$ \\
        Calculate low-variance threshold: $P_{t_\mathtt{var}} \gets \textbf{Quantile}(A_\mathtt{var}, t_\mathtt{var})$ \\
        Generate low-variance mask: $M_\mathtt{var} \gets \textbf{Mask}(A_\mathtt{var}, P_{t_\mathtt{var}})$ \\
        \textbf{Store} $M_\mathtt{var}$ for applying adversarial noise
    }
    \Else{
        \tcp{\textcolor{blue}{Compute Adversarial Loss}}
        Compute adversarial attention map: $A_\mathtt{map}' \gets \textbf{Softmax}(Q_x K_{(x+\delta)}^T / \sqrt{d})$ \\
        Compute variance: $A_\mathtt{var}' \gets \textbf{Var}(A_\mathtt{map}')$ \\
        Calculate attention loss in low-variance regions: $\mathcal{L}_\mathtt{attn} \gets \mathcal{L}_\mathtt{attn} + \mathcal{F}(\Delta)$, \newline
        where $\Delta = (\sigma_\mathtt{max} - A_\mathtt{var}') \odot M_\mathtt{var}$
    }
    \textbf{Subsequent steps are not shown here.}
\end{algorithm}

\subsection{MTCNN Attack}
\label{sup:a2}
\myparagraph{Model architecture.}

\begin{figure}[ht]
\vspace{-5pt}
\centering
\includegraphics[width=\textwidth]{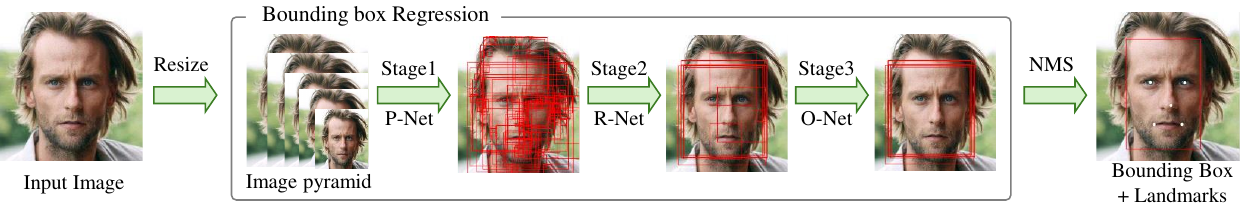}
\captionof{figure}
{\textbf{MTCNN model architecture overview}. }
\label{fig:MTCNN_arc}
\end{figure}

The Multi-task Cascaded Convolutional Neural Network (MTCNN) is a deep learning-based framework for face detection and facial landmark localization. Its architecture consists of three cascaded convolutional neural networks, each refining face candidates while ensuring computational efficiency (see Fig.\ref{fig:MTCNN_arc}). The Proposal Network (P-Net) employs a sliding window to scan the input image, generating bounding box proposals and associated confidence scores. Non-maximum suppression (NMS) is applied to remove redundant proposals. The Refine Network (R-Net) filters the bounding boxes further, reducing false positives and improving localization accuracy. Finally, the Output Network (O-Net) refines the bounding boxes and predicts precise facial landmark locations for face alignment. A key strength of MTCNN lies in its multi-scale input processing strategy. By resizing the input image across multiple scales, the network effectively captures faces of varying sizes, ensuring robust detection under diverse scenarios. This approach enables the P-Net to detect both large and small faces within a single pipeline, generating a comprehensive set of bounding box proposals. The cascaded structure leverages these multi-scale candidates, progressively refining them to achieve high detection accuracy and precision, even in complex scenes with occlusions or extreme pose variations.

\clearpage
\begin{figure}[ht]
\centering
\includegraphics[width=\textwidth]{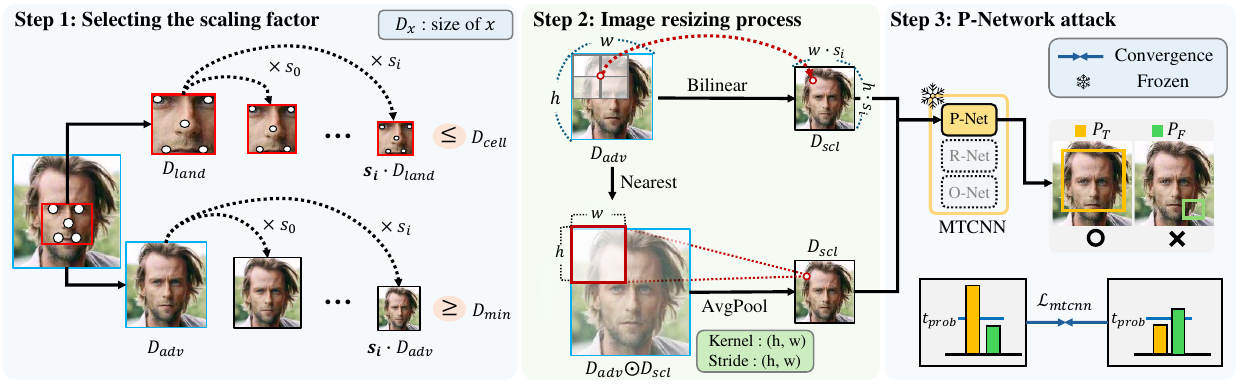}
\captionof{figure}
{\textbf{MTCNN Attack Overview}. The attack process on MTCNN consists of three parts: (i) Selecting the scaling factor $s_i$, where the scale value is chosen according to Eq.\ref{method_eq:14}; (ii) Image resizing process, where we extend the robustness of resizing modes by using both Bilinear interpolation and our proposed Area-based method; (iii) P-Net attack, which decreases the probability values of candidate scales.}
\label{fig:supple_mtcnn}
\end{figure}

\myparagraph{Details of the scaling factor selection process.}
MTCNN uses a multi-scale approach for face detection, which motivates us to extend the robustness of our adversarial noise across different scaling factors. To achieve this, we calculate the loss over multiple scales by dividing the image into several scales (see Fig.\ref{fig:supple_mtcnn}).
The process of selecting the optimal scaling factor is as follows: Initially, we calculate the minimum bounding box size $D_\mathtt{land}$ that encompasses key facial landmarks (eyes, nose, mouth) in the input image, obtained by passing the original image through MTCNN.
Suitable scale values $s_i$ are chosen to adjust the initial bounding box size $D_\mathtt{cell}$ to be larger than $D_\mathtt{land}$, while ensuring that the scaled input image size $D_\mathtt{adv}$ remains greater than the minimum allowable size $D_\mathtt{min}$. This is mathematically expressed as:
\begin{align}\label{method_eq:14}
    Scales = \left\{ s_i \mathrel{\Bigg|} {s_i} \cdot D_\mathtt{land} \leq D_\mathtt{cell}, \ s_i \cdot D_\mathtt{adv} \geq D_\mathtt{min} \right\}
\end{align}
where $s_i$ is defined as $s_i = \frac{D_\mathtt{cell}}{D_\mathtt{min}} \times k^{i-1}$, with $k$ being a predefined scale factor and $i$ a non-negative integer.  
This ensures that only bounding boxes reaching MTCNN's final layers are effectively targeted. 

\myparagraph{Algorithm.}
The image resizing process and the P-Network attack method are summarized in Algorithm~\ref{alg:mtcnn}.
\begin{algorithm}[ht]
    \setlength{\belowcaptionskip}{-20pt}
    \caption{Adversarial loss in MTCNN Attack.}
    \label{alg:mtcnn}
    \Indm
    \KwIn{
        source face image $x$, perturbation $\delta$, probability threshold $t_\mathtt{prob}$, image resize scale set $Scales$, mtcnn P-Network $\mathcal{T}$, mtcnn loss $\mathcal{L}_\mathtt{mtcnn}$, mtcnn loss function $\mathcal{F}$
    }
    \KwResult{
        added mtcnn loss $\mathcal{L}_\mathtt{mtcnn}$
    }
    \Indp
    Update input image with perturbation: $x_\mathtt{adv} \gets x + \delta$ \\
    Get input image size: $D_\mathtt{adv} \gets Shape(x_\mathtt{adv})$ \\
    Set kernel and stride sizes: $K, S \gets D_\mathtt{adv}$ \\
    \For{$s_i$ in $Scales$}{
        Set scaled image size: $D_\mathtt{scl} \gets s_i \times D_\mathtt{adv}$ \\
        Compute intermediate image size: $D_\mathtt{int} \gets D_\mathtt{adv} \odot D_\mathtt{scl}$ \\
        Upscaling image by using $\mathtt{NEAREST}$: $\hat{x}_\mathtt{adv} \gets \textbf{Scale}(x_\mathtt{adv}, D_\mathtt{int})$ \\
        Apply average pooling: $\tilde{x}_\mathtt{adv} \gets \textbf{Pool}(\hat{x}_\mathtt{adv}, K, S)$ \\
        Obtain bbox probability: $P_\mathtt{T}, P_\mathtt{F} \gets \mathcal{T}(\tilde{x}_\mathtt{adv})$ \\
        Generate high-probability mask: $M_\mathtt{prob} \gets \textbf{Mask}(P_\mathtt{T}, t_\mathtt{prob})$ \\
        Calculate mtcnn loss in mask region: $\mathcal{L}_\mathtt{mtcnn} \gets \mathcal{L}_\mathtt{mtcnn} + \mathcal{F}(\Delta)$, \newline
        where $\Delta = (\mathcal{T}(\tilde{x}_\mathtt{adv}) - p_\mathtt{gt}) \odot M_\mathtt{prob}$
    }
\end{algorithm}

\clearpage
\section{Additional Related Work}
\label{sup:b}
\myparagraph{Deepfake adversarial attack.}
Existing research on adversarial attacks against deepfakes has focused on two main approaches: one involves targeting deepfake models based on the structural properties of specific GANs, and the other focuses on facial feature extractors to attack multiple deepfake models that use them.
Studies such as \cite{ruiz2020disrupting, huang2022cmua, wang2022anti} have focused on degrading the quality of images by targeting various GANs~\cite{zhu2017unpaired, choi2018stargan, tang2019attention, he2019attgan, li2021image}. However, these approaches are ineffective against DM-based models~\cite{kim2212diffface, zhao2023diffswap, wang2024face}.
As a study that attacks facial feature extractors, \cite{li2024landmarkbreaker} performs adversarial attacks on several face landmark models~\cite{qian2019aggregation, yin2020fan, wang2020deep}, although the extractors targeted in this study are now less commonly used.
\cite{kaziakhmedov2019real} disrupt face detection targeting the MTCNN model by applying specific patches, but this approach has the limitation of being visible. Another method attacking the same model, \cite{zhang2022multi}, propose using $\mathtt{BILINEAR}$ interpolation to attack across multiple scales. However, since the $\mathtt{BILINEAR}$ mode only uses specific anchor points during interpolation, adversarial noise generated with this approach easily loses effectiveness when other interpolation modes are applied.

\myparagraph{Diffusion adversarial attack.}
As image editing techniques utilizing DMs have gained traction, research on adversarial attacks targeting these architectures has progressed significantly.
AdvDM~\cite{liang2023adversarial} generates adversarial examples by optimizing latent variables sampled from the reverse process of a DM.
Similarly, Glaze~\cite{shan2023glaze} investigates the latent space, generating adversarial noise and proposing a noise clamping technique based on $\mathtt{LPIPS}$ minimizing perceptual distortion of the original image.
Photoguard~\cite{salman2023raising} is noteworthy for introducing the concept of encoder attacks, and separately, it presents a diffusion attack that utilizes the denoised generated image.
Mist~\cite{liang2023mist} combines the semantic loss proposed in \cite{liang2023adversarial} with the textual loss from \cite{salman2023raising}, leading to a novel loss function that enables the generation of transferable adversarial examples against various diffusion-based attacks. 
Diff-Protect~\cite{xue2023toward} proposes a novel approach that updates by minimizing loss, unlike previous studies.
DiffusionGuard~\cite{choi2024diffusionguard} introduces adversarial noise early in the diffusion process, preventing image editing techniques from reproducing sensitive areas.
All previous research has been directed toward protecting images when they are utilized directly in DMs, as depicted in Fig. 2(a) in the main paper.

\myparagraph{Adversarial noise with frequency-domain.}
There are various approaches utilizing frequency in generating adversarial noise. Maiya et al.~\cite{maiya2021frequency} suggested that using frequency is effective in designing imperceptible noise while Wang et al.~\cite{wang2020high} argued that high-frequency components are effective for attacking CNN-based models. On the other hand, recent studies \cite{sharma2019effectiveness, guo2018low} has demonstrated that it is possible to attack DNN-based models~\cite{moosavi2016deepfool, sharif2016accessorize} effectively using only low-frequency components. Additionally, AdvDrop~\cite{duan2021advdrop} showed that transformations in the frequency domain of images can induce misclassification. Ling et al.~\cite{ling2024fdt} proposed the frequency data transformation(FDT) method to improve transferability between models in black-box attacks.

\section{Additional Experimental Details}
\subsection{Implementation Details}
\label{sup:c1}
In this paper, we generate \textit{FaceShield} by utilizing the mid-layer cross-attention of the open-source Stable Diffusion Model v1.5~\cite{Rombach_2022_CVPR}, the upper part of the CLIP Image Projector in the CLIP Model~\cite{radford2021learning}, only the P-Network from the PyTorch version of MTCNN~\cite{zhang2016joint}, and two variants of ArcFace~\cite{deng2019arcface}.
All images are resized to $512 \times 512$ before processing, and experiments are conducted on an RTX A6000.
A more detailed description is provided in Table~\ref{table:hyper}, where the same hyperparameters are applied as in the baseline methods~\cite{liang2023adversarial,liang2023mist,salman2023raising,xue2023toward} for generating noise.
\begin{table}[ht]
\centering
\adjustbox{max width=0.45\textwidth}{
    \begin{tabular}{cccc}
    \hlineB{2.5}
    \textbf{Norm} & \textbf{$\epsilon$} & \textbf{step size} & \textbf{number of steps} \\ 
    \hline
    $\ell_\infty$ & 12/255 & 1/255 & 30 \\
    \hlineB{2.5}
    \end{tabular}}
\caption{Hyperparameters used for the PGD attacks.}
\label{table:hyper}
\end{table}

\noindent As a result, \textit{FaceShield} achieves 24 seconds per image with only 15 GB of memory, demonstrating significantly lower resource costs compared to baseline methods, as shown in Table~\ref{table:resource}. 
This efficiency is achieved through three key optimizations:
(i) Restricting the input to the Conditioned Face Attack (CFA) module, ensuring the process focuses solely on facial regions.
(ii) Extracting gradients from the condition path (Fig.3 in the main paper), eliminating the need for gradient accumulation across multiple timesteps.
(iii) Updating only the mid-layer of the UNet, rather than optimizing the entire network.
These optimizations enable \textit{FaceShield} to achieve high performance with minimal computational resources.
\begin{table}[ht]
\centering
\adjustbox{max width=0.45\textwidth}{
    \begin{tabular}{c|c|c|r|r}
    \hlineB{2.5}
    \textbf{Baseline} & ISM$\downarrow$ & LPIPS$\downarrow$ & VRAM & Sec.$\downarrow$\\
    \hline
    \text{AdvDM~\cite{liang2023adversarial}} & 0.288 & 0.4214 & 20 GB & 39 \\
    \text{Mist~\cite{liang2023mist}} & 0.291 & 0.5492 & 22 GB & 80 \\
    \text{PhotoGuard~\cite{salman2023raising}} & 0.294 & 0.5515 & 28 GB & 234 \\
    \text{SDST~\cite{xue2023toward}} & 0.303 & 0.5409 & \textbf{11 GB} & 34 \\
    \hline
    \text{\textbf{Ours}} & \textbf{0.168} & \textbf{0.2017} & \underline{15 GB} & \textbf{24} \\ 
    \hlineB{2.5}
    \end{tabular}}
\caption{Comparison of resource costs with baseline methods.}
\label{table:resource}
\end{table}

\myparagraph{Gaussian Blur.}
To achieve more precise detection of intensity variations between adjacent pixels, we employ a $3 \times 3$ Sobel matrix. Its compact size ensures faster convolution operations and reduces memory consumption, which is crucial for iterative computations. Subsequently, a $9 \times 9$ padding is applied to the detected regions to generate thicker masks, ensuring smoother transitions during the subsequent Gaussian blur step and mitigating abrupt changes.

\begin{figure}[ht]
\centering
\includegraphics[width=0.5\columnwidth]{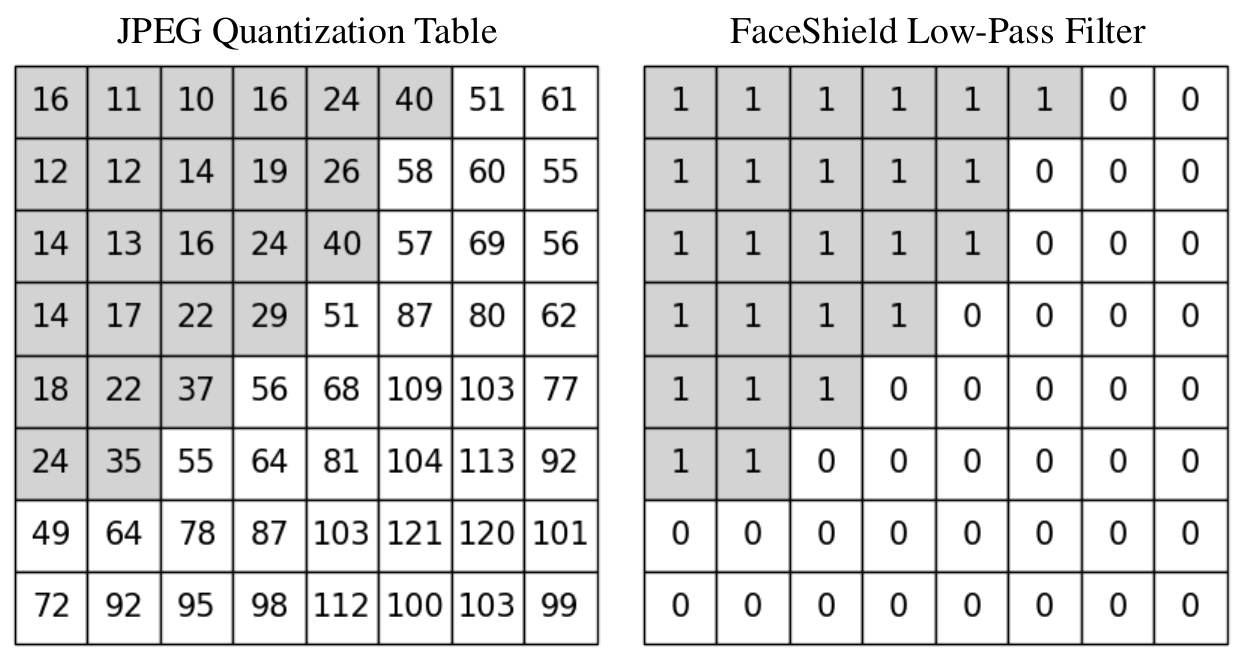}
\captionof{figure}
{The table on the left shows the Luminance Quantization Table used in the JPEG compression process. The table on the right illustrates the \textit{FaceShield}'s Low-pass Filter, which is created by selecting only the values below 40.}
\label{fig:supple_lpf}
\end{figure}

\myparagraph{Low-pass Filter.}
We utilize perturbations in the frequency domain by performing an 8×8 patch division followed by a Discrete Cosine Transform (DCT). This design is inspired by the JPEG compression scheme, which operates on 8×8 blocks and employs a Quantization Table to prioritize low-frequency components. Furthermore, the 8×8 patch division offers computational efficiency advantages compared to approaches without such division during the DCT process.
Unlike JPEG compression, we skip the RGB-to-YCbCr color space transformation. This decision is based on two considerations: (i) perturbations inherently contain both positive and negative values, which are incompatible with the typical range constraints of the YCbCr domain, and (ii) experiments demonstrate that handling frequencies directly in the RGB domain is sufficient to achieve our performance objectives without compromising effectiveness.
The coefficients for our low-pass filter are selected from the Luminance Quantization Table, focusing exclusively on values below 40, as illustrated in Fig.\ref{fig:supple_lpf}.

\subsection{Human Evaluation}
\label{sup:c2}
\begin{figure}[ht]
\centering
\includegraphics[width=\columnwidth]{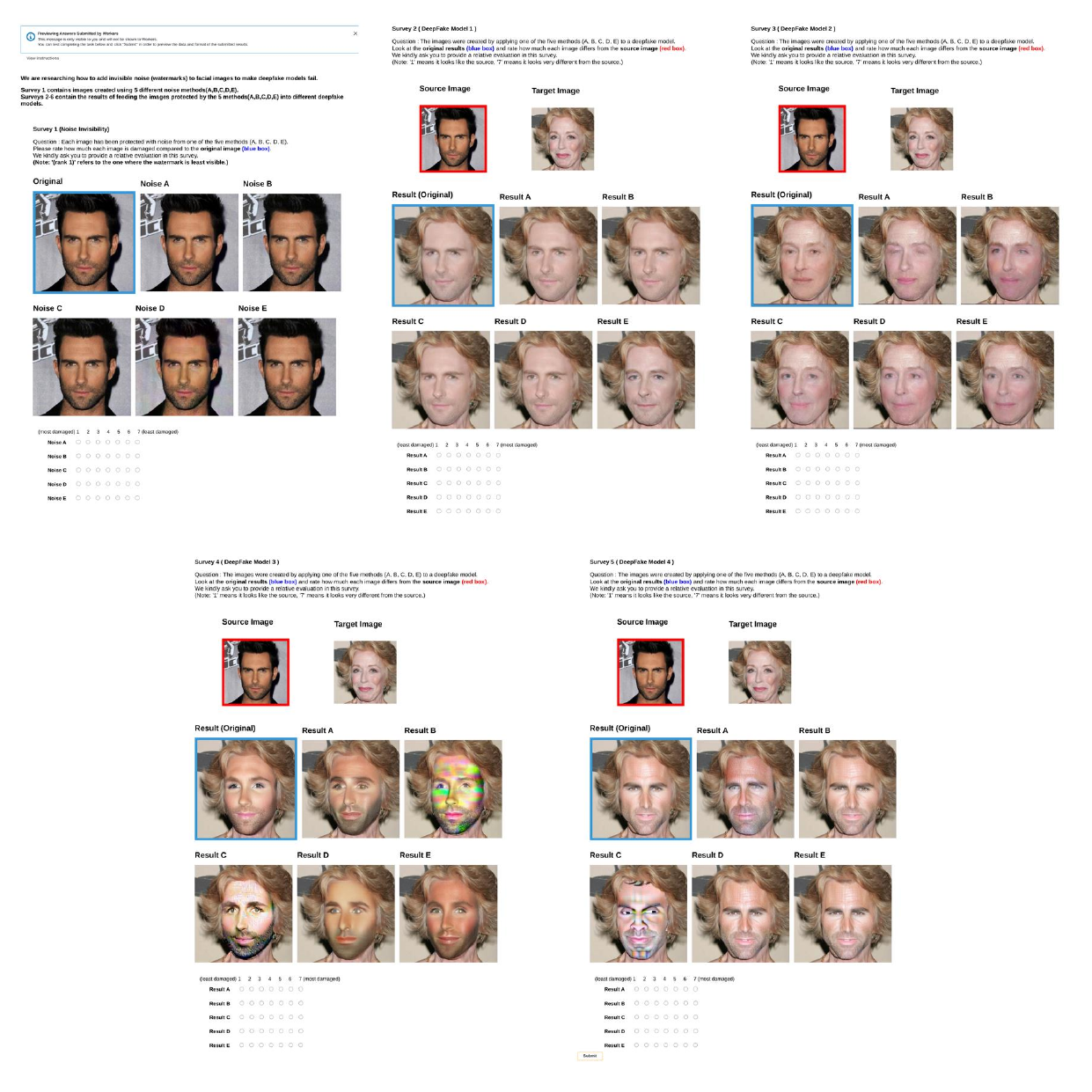}
\captionof{figure}
{\textbf{Human Evaluation Survey}. Survey 1 (the first figure) evaluates the visibility of the noise, while Surveys 2-5 (the remaining figures) assess the protection performance across different deepfake models~\cite{kim2212diffface,zhao2023diffswap,wang2024face,ye2023ip}. The scoring scale ranges from 1 to 7, and to ensure fairness, the placement of comparison methods was randomly shuffled for each survey.
}
\label{fig:supple_he}
\end{figure}

We conduct a human evaluation study to assess the visibility of the noise and the protection performance across four deepfake models~\cite{kim2212diffface,zhao2023diffswap,wang2024face,ye2023ip}, along with four baseline methods~\cite{liang2023adversarial,liang2023mist,salman2023raising,xue2023toward}. Specifically, participants are asked to score images on a scale from 1 (low performance) to 7 (high performance) in response to the following two questions: (i) \textit{"How much each image is damaged compared to the original image?"}, which measures the visibility of the protective noise pattern relative to each baseline method, and (ii) \textit{"How much each image differs from the source image?"}, which evaluates how effectively each method prevents the deepfake models from reflecting the original source face. 
We use 20 images (10 from the CelebA-HQ dataset and 10 from the VGGFace2-HQ dataset) across four deepfake models, with 100 participants providing their ratings. To enhance fairness, the positions of the compared methods within each question are randomly shuffled.
An example survey is shown in Fig.\ref{fig:supple_he}.

\clearpage
\section{Additional Ablation Study}
\subsection{MTCNN Resize Robustness}
\label{sup:d1}
\begin{figure}[ht]
    \centering
    \includegraphics[width=\columnwidth]{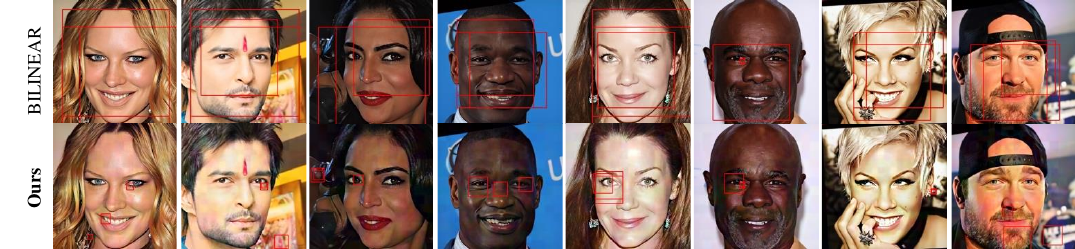}
    \caption{We compare the performance of the image resize method using only $\mathtt{BILINEAR}$ interpolation (top) and our proposed approach (bottom). Experiments are conducted with the default MTCNN resizing mode, $\mathtt{CV2.INTER\_AREA}$ . The bounding boxes (red boxes) shown represent the top three outputs from the P-Net with the highest confidence scores.}
    \label{fig:supple_bbox}
\end{figure}

The experimental results for MTCNN, as discussed in the Ablation Study of the main paper, are presented through both quantitative and qualitative evaluations.
Specifically, Table~\ref{table:mtcnn_opencv} and Table~\ref{table:mtcnn_pil} provide quantitative metrics, while Fig.\ref{fig:supple_bbox} illustrates how the detected regions propagate to the subsequent network when face detection fails at the P-Network stage.
These results demonstrate the superiority of the newly proposed method in \textit{FaceShield} compared to the $\mathtt{BILINEAR}$ approach introduced in prior work~\cite{zhang2022multi}, which aimed to perturb the MTCNN model. In particular, Table~\ref{table:mtcnn_opencv} evaluates various scaling modes provided by $\mathtt{OpenCV}$, while Table~\ref{table:mtcnn_pil} focuses on those offered by $\mathtt{Pillow}$. The experiments were conducted using both the PyTorch and TensorFlow versions of the framework.
For comprehensive evaluation, we utilized 3,000 images each from the CelebA-HQ~\cite{karras2017progressive} and VGGFace2-HQ~\cite{simswapplusplus} datasets. The results confirm that \textit{FaceShield} achieves superior coverage across diverse scaling modes compared to previous approaches.

\begin{table}[ht]
\centering
\adjustbox{max width=\columnwidth}{
\begin{tabular}{c|cccccc}
    \hlineB{2.5}
    \textbf{Dataset} & \multicolumn{6}{c}{CelebA-HQ~\cite{karras2017progressive}} \\
    \hline
    \textbf{Method} & $\mathtt{BILINEAR}$ & $\mathtt{AREA}$ & $\mathtt{NEAREST}$ & $\mathtt{CUBIC}$ & $\mathtt{LANC}$ & $\mathtt{EXACT}$ \\
    \hline
    BILINEAR & 93.77\% & 0.07\% & 0.40\% & 95.73\% & 95.67\% & 93.77\% \\
    \cellcolor{gray!30}\textbf{Ours} & \cellcolor{gray!30}97.31\% & \cellcolor{gray!30}94.17\% & \cellcolor{gray!30}4.13\% & \cellcolor{gray!30}97.10\% & \cellcolor{gray!30}97.00\% & \cellcolor{gray!30}97.30\% \\
    \hline
    \hline
    \textbf{Dataset} & \multicolumn{6}{c}{VGGFace2-HQ~\cite{simswapplusplus}} \\
    \hline
    \textbf{Method} & $\mathtt{BILINEAR}$ & $\mathtt{AREA}$ & $\mathtt{NEAREST}$ & $\mathtt{CUBIC}$ & $\mathtt{LANC}$ & $\mathtt{EXACT}$ \\
    \hline
    BILINEAR & 87.23\% & 0.17\% & 0.37\% & 94.63\% & 94.43\% & 88.93\% \\
    \cellcolor{gray!30}\textbf{Ours} & \cellcolor{gray!30}89.20\% & \cellcolor{gray!30}72.93\% & \cellcolor{gray!30}2.47\% & \cellcolor{gray!30}94.93\% & \cellcolor{gray!30}95.27\% & \cellcolor{gray!30}89.33\% \\
    \hlineB{2.5}
    \end{tabular}
}
\caption{The metric values represent the detection failure rates of the MTCNN~\cite{zhang2016joint} model. Our scaling method demonstrates greater robustness across various scaling modes in the $\mathtt{OpenCV}$ Library compared to the existing approach, with particularly notable performance in the model's default setting, $\mathtt{AREA}$.}
\vspace{-5pt}
\label{table:mtcnn_opencv}
\end{table}

\begin{table}[ht]
\centering
\adjustbox{max width=0.7\columnwidth}{
\begin{tabular}{c|cccccc}
    \hlineB{2.5}
    \textbf{Dataset} & \multicolumn{6}{c}{CelebA-HQ~\cite{karras2017progressive}} \\
    \hline
    \textbf{Method} & $\mathtt{BILINEAR}$ & $\mathtt{BOX}$ & $\mathtt{NEAREST}$ & $\mathtt{BICUBIC}$ & $\mathtt{LANCZ0S}$ & $\mathtt{HAMMING}$ \\
    \hline
    BILINEAR & 0.70\% & 0.80\% & 79.73\% & 0.57\% & 0.84\% & 0.70\% \\
    \rowcolor{gray!30}
    \textbf{Ours} & 10.67\% & 98.57\% & 97.90\% & 16.90\% & 16.30\% & 37.53\% \\
    \hline
    \hline
    \textbf{Dataset} & \multicolumn{6}{c}{VGGFace2-HQ~\cite{simswapplusplus}} \\
    \hline
    \textbf{Method} & $\mathtt{BILINEAR}$ & $\mathtt{BOX}$ & $\mathtt{NEAREST}$ & $\mathtt{BICUBIC}$ & $\mathtt{LANCZ0S}$ & $\mathtt{HAMMING}$ \\
    \hline
    BILINEAR & 1.37\% & 1.87\% & 68.83\% & 1.47\% & 1.60\% & 1.63\% \\
    \rowcolor{gray!30}
    \textbf{Ours} & 12.83\% & 84.20\% & 87.97\% & 16.03\% & 15.53\% & 28.53\% \\
    \hlineB{2.5}
    \end{tabular}
}
\caption{The metric values represent the detection failure rates of the MTCNN~\cite{zhang2016joint} model. Our scaling method demonstrates greater robustness across various scaling modes in the $\mathtt{Pillow}$ Library compared to the existing approach.}
\label{table:mtcnn_pil}
\end{table}

\clearpage
\subsection{Gaussian blur Effect}
\label{sup:d2}
The qualitative results of the Gaussian blur effect, mentioned in the Ablation Study of the main paper, are presented in the following Fig.\ref{fig:blur}, comparing the cases with and without its application. As shown in the figure on the right, Sobel filtering is applied to achieve effective invisibility while maintaining maximum performance, resulting in blurred areas where noticeable differences between adjacent regions exist. Additional examples of the results are provided in Fig.\ref{fig:supp_gaussian_blur}.
\begin{figure}[h]
    \centering
    \includegraphics[width=0.6\columnwidth]{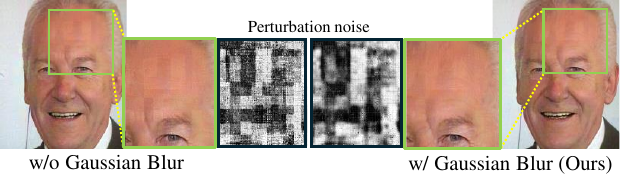}
    \caption{By detecting regions with large intensity differences between adjacent RGB pixels in the perturbation, a blur effect is applied, enhancing the invisibility of the noise.}
    \label{fig:blur}
\end{figure}

\begin{figure}[ht]
    \centering
    \includegraphics[width=0.8\columnwidth]{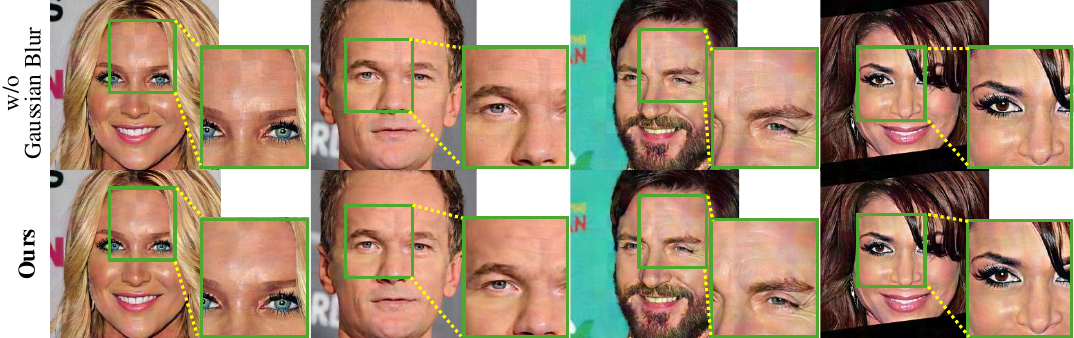}
    \caption{Qualitative comparison between the case with Gaussian Blur (bottom) and without Gaussian Blur (top).}
    \label{fig:supp_gaussian_blur}
\end{figure}

\clearpage
\section{Evaluating FaceShield under Image Purifications}
\label{sup:e}
We conduct additional experiments to demonstrate the robustness of \textit{FaceShield} leveraging low-frequency components against various image purification techniques. Specifically, we evaluate the performance under three primary scenarios.

\begin{figure}[ht]
    \centering
    \includegraphics[width=0.8\columnwidth]{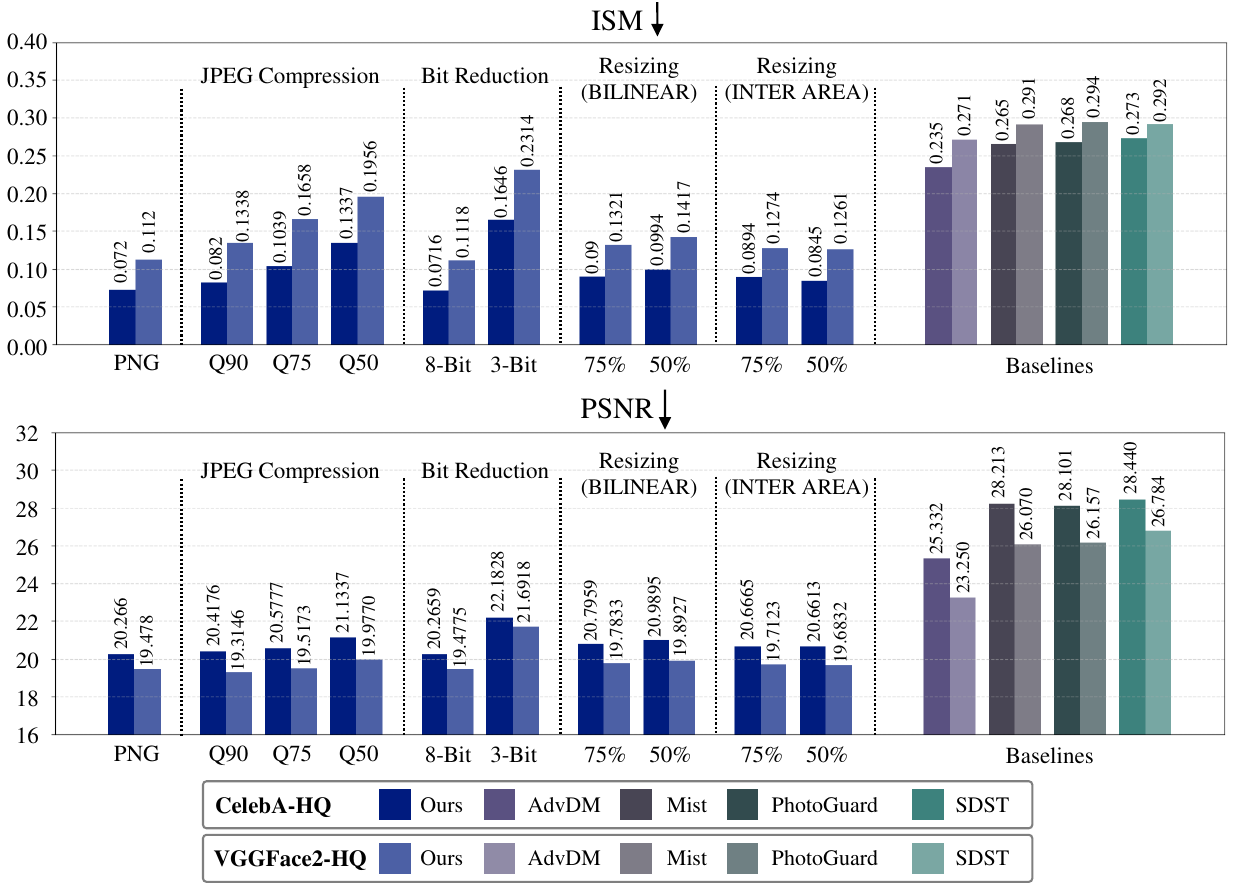}
    \caption{Quantitative results of \textit{FaceShield}-protected images after passing through various purification methods and evaluated on a deepfake model~\cite{ye2023ip}.
    Our method demonstrates robustness against various purification methods, including JPEG compression, bit reduction, and two types of resizing, with its performance compared to baseline methods~\cite{liang2023adversarial,liang2023mist,salman2023raising,xue2023toward}.
    The results, measured using PSNR and Identity Score Matching (ISM), show that our method closely resembles lossless (PNG) outcomes while consistently outperforming the baselines. Both metrics indicate better performance with lower values.}
    \label{fig:supp_graph}
\end{figure}

\begin{figure}[ht]
    \centering
    \includegraphics[width=0.9\columnwidth]{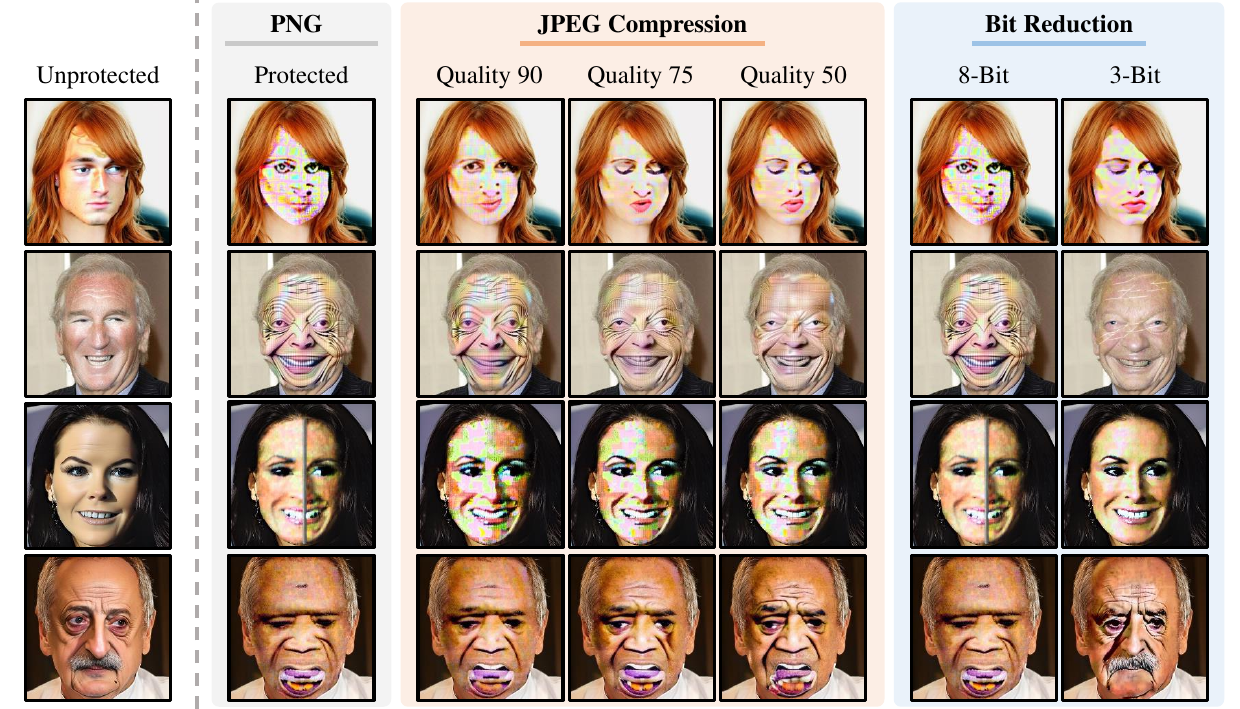}
    \caption{The results of applying three levels of JPEG compression and two levels of bit reduction to images protected by \textit{FaceShield}, followed by evaluation on a deepfake model~\cite{ye2023ip}, show that the performance degradation is minimal compared to lossless storage (PNG).}
    \label{fig:sup_jpeg_bit}
\end{figure}

\begin{figure}[ht]
    \centering
    \includegraphics[width=0.9\columnwidth]{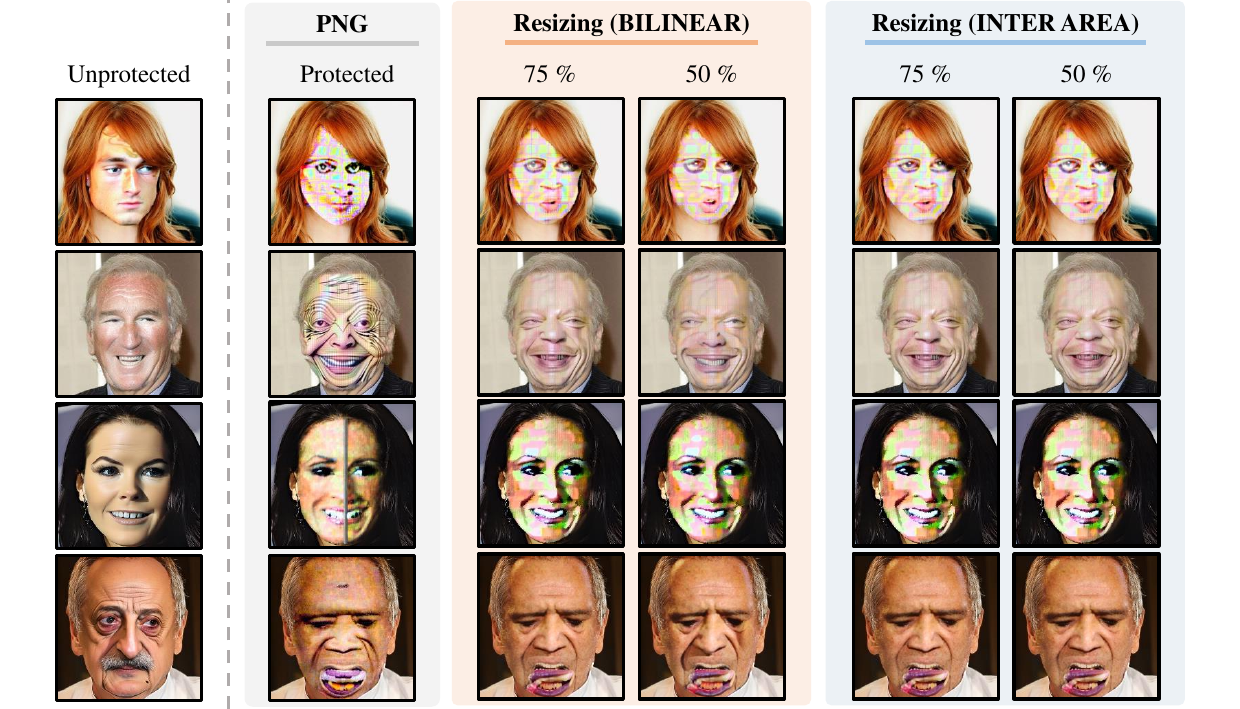}
    \caption{The results of applying two types of resizing methods, with 75\% and 50\% scaling, to images protected by \textit{FaceShield}, followed by evaluation on a deepfake model~\cite{ye2023ip}, show that the performance degradation is minimal compared to lossless storage (PNG).}
    \label{fig:sup_resize}
\end{figure}

\begin{itemize}
    \item \textbf{JPEG compression}:  Images are compressed at quality levels of 90, 75, and 50 to introduce distortions.
    \item \textbf{Bit reduction}: Images are quantized to 8-Bit and 3-Bit formats, simulating lossy storage conditions.
    \item \textbf{Resizing}: Images are resized to 75\% and 50\% of their original dimensions and then restored to their original size. Two interpolation methods, $\mathtt{BILINEAR}$ and $\mathtt{INTER\_AREA}$, are applied during resizing.
\end{itemize}
These experiments are conducted using the IP-Adapter model~\cite{ye2023ip}, with the same dataset as in Table 1 in the main paper.
The quantitative results for ISM and PSNR are presented in Fig.\ref{fig:supp_graph}, while the qualitative results are shown in Fig.\ref{fig:sup_jpeg_bit} and Fig.\ref{fig:sup_resize}.
As shown in the results, \textit{FaceShield} causes only minor performance degradation across various purification methods, yet still demonstrates superior performance compared to other baselines~\cite{liang2023adversarial,liang2023mist,salman2023raising,xue2023toward}, proving its remarkable robustness.

\section{Additional Qualitative Results}
\label{sup:f}
\begin{figure}[ht]
    \centering
    \includegraphics[width=0.9\columnwidth]{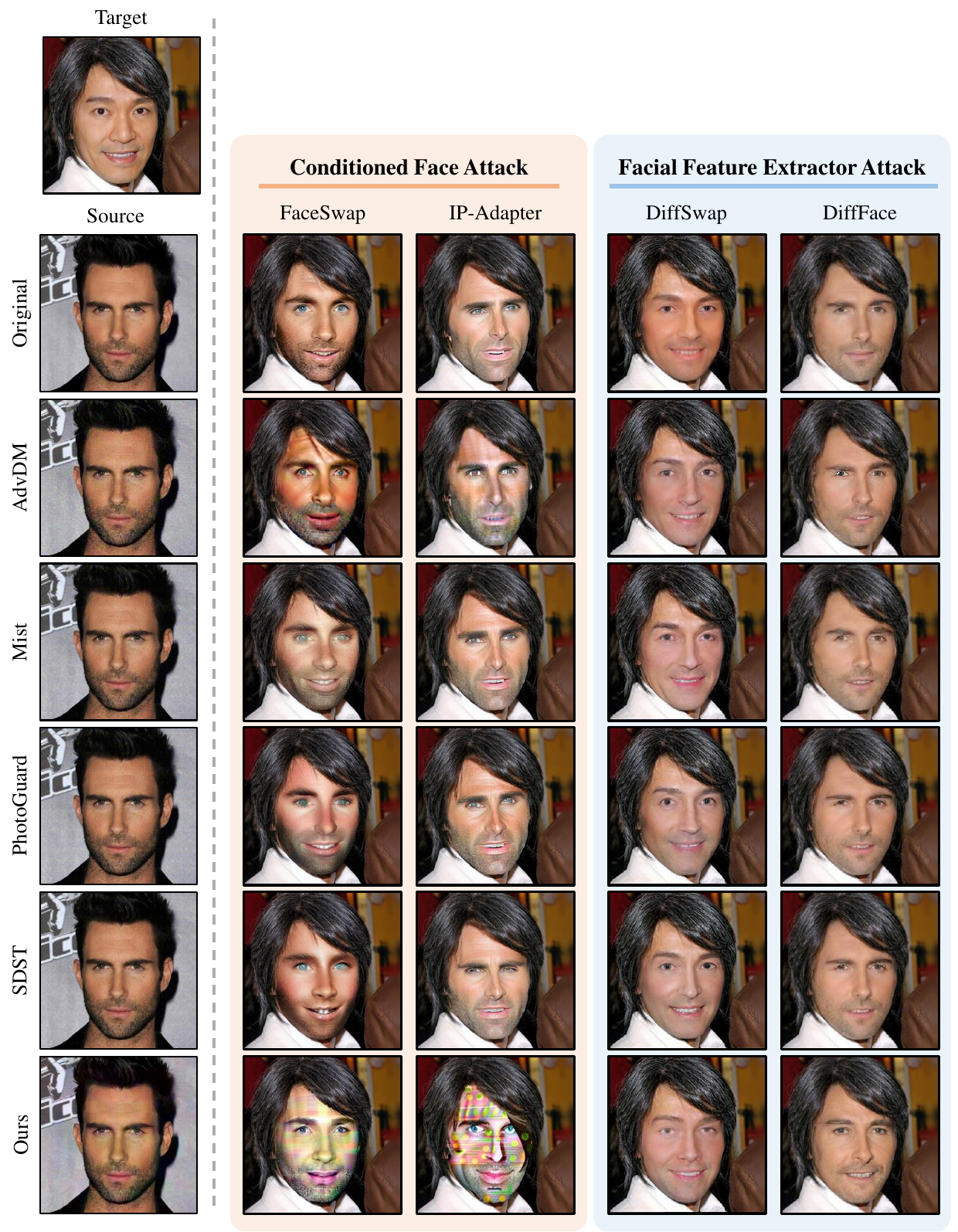}
    \caption{Qualitative comparisons with AdvDM~\cite{liang2023adversarial}, Mist~\cite{liang2023mist}, PhotoGuard~\cite{salman2023raising}, and SDST~\cite{xue2023toward} across four diffusion-based deepfake models: FaceSwap~\cite{wang2024face}, IP-Adapter~\cite{ye2023ip}, DiffSwap~\cite{zhao2023diffswap}, and DiffFace~\cite{kim2212diffface}.}
    \label{fig:supple_fig_full_1}
\end{figure}

\begin{figure}[ht]
    \centering
    \includegraphics[width=0.9\columnwidth]{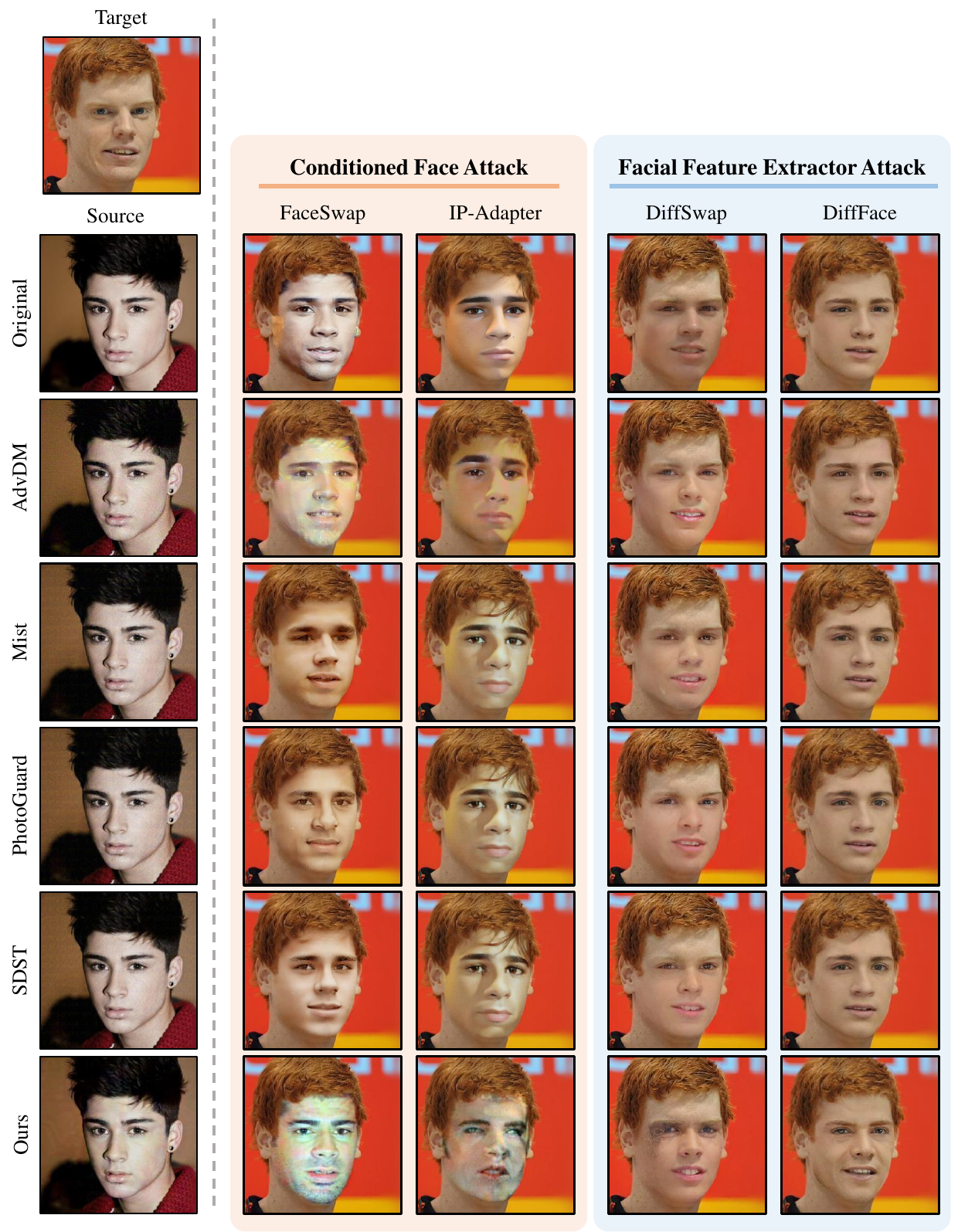}
    \caption{Qualitative comparisons with AdvDM~\cite{liang2023adversarial}, Mist~\cite{liang2023mist}, PhotoGuard~\cite{salman2023raising}, and SDST~\cite{xue2023toward} across four diffusion-based deepfake models: FaceSwap~\cite{wang2024face}, IP-Adapter~\cite{ye2023ip}, DiffSwap~\cite{zhao2023diffswap}, and DiffFace~\cite{kim2212diffface}.}
    \label{fig:supple_fig_full_2}
\end{figure}

\begin{figure}[ht]
    \centering
    \includegraphics[width=0.9\columnwidth]{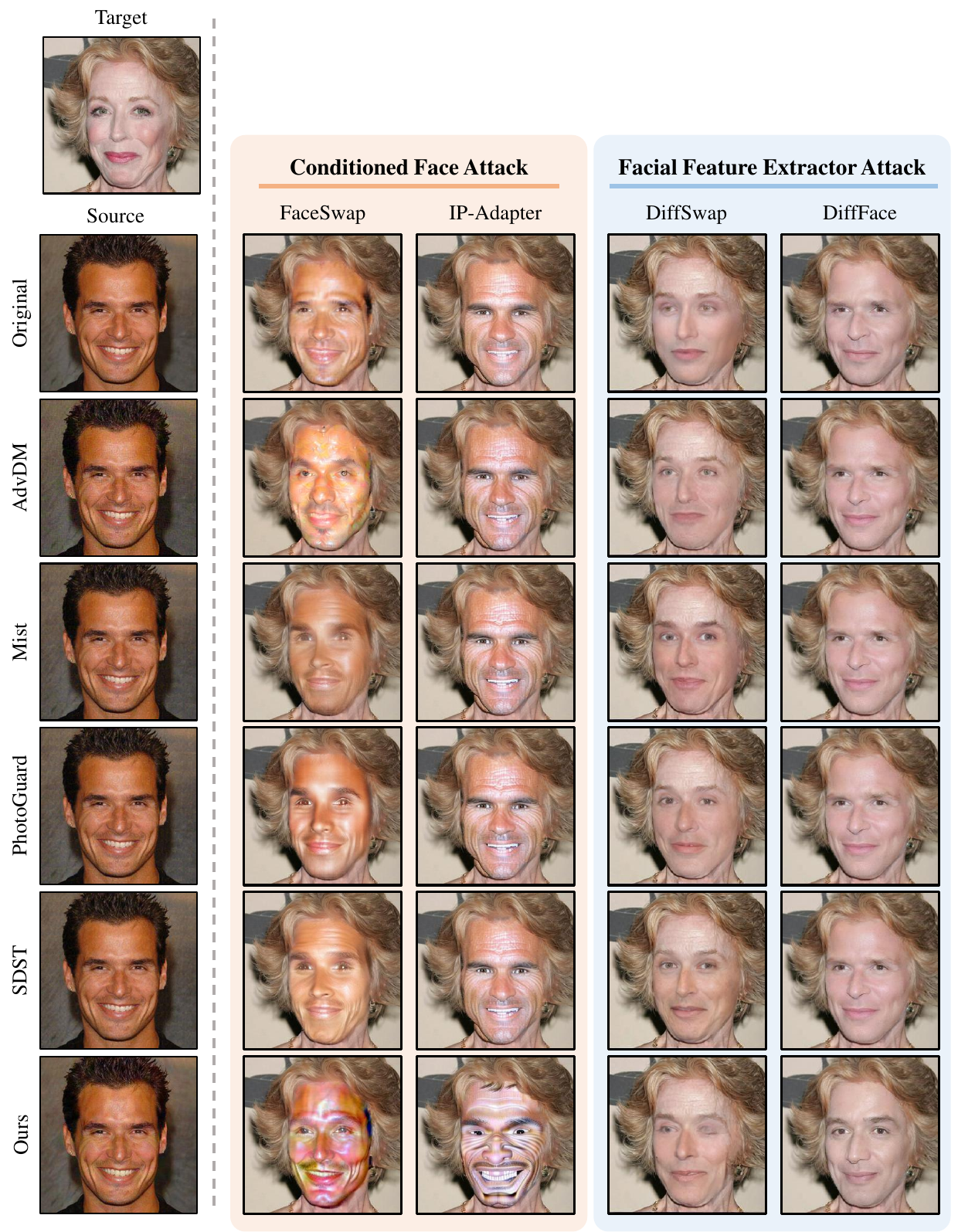}
    \caption{Qualitative comparisons with AdvDM~\cite{liang2023adversarial}, Mist~\cite{liang2023mist}, PhotoGuard~\cite{salman2023raising}, and SDST~\cite{xue2023toward} across four diffusion-based deepfake models: FaceSwap~\cite{wang2024face}, IP-Adapter~\cite{ye2023ip}, DiffSwap~\cite{zhao2023diffswap}, and DiffFace~\cite{kim2212diffface}.}
    \label{fig:supple_fig_full_3}
\end{figure}

\begin{figure}[ht]
    \centering
    \includegraphics[width=0.9\columnwidth]{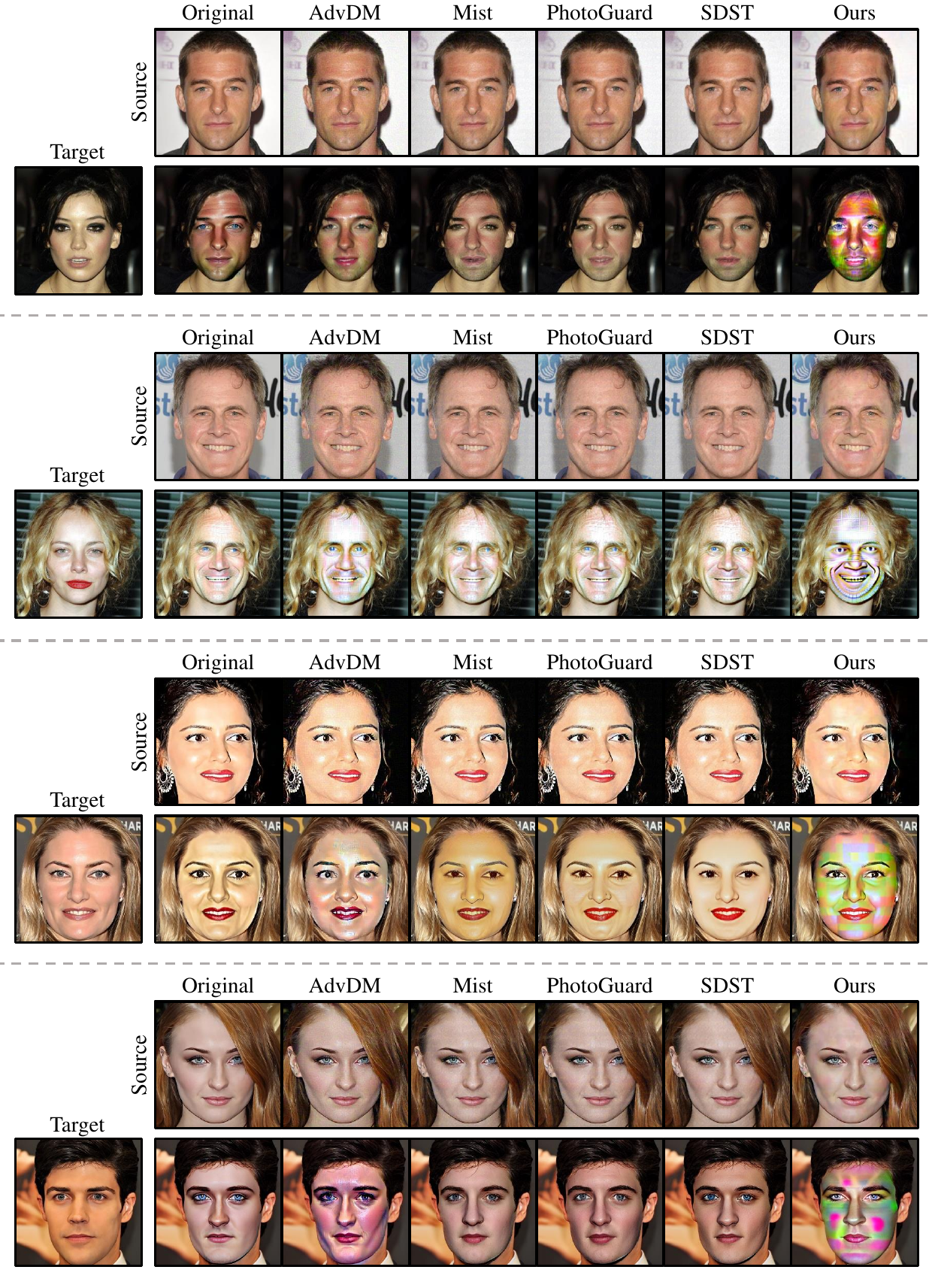}
    \caption{Qualitative comparisons for \textbf{FaceSwap}~\cite{wang2024face}.}
    \label{fig:supple_faceswap}
\end{figure}

\begin{figure}[ht]
    \centering
    \includegraphics[width=0.9\columnwidth]{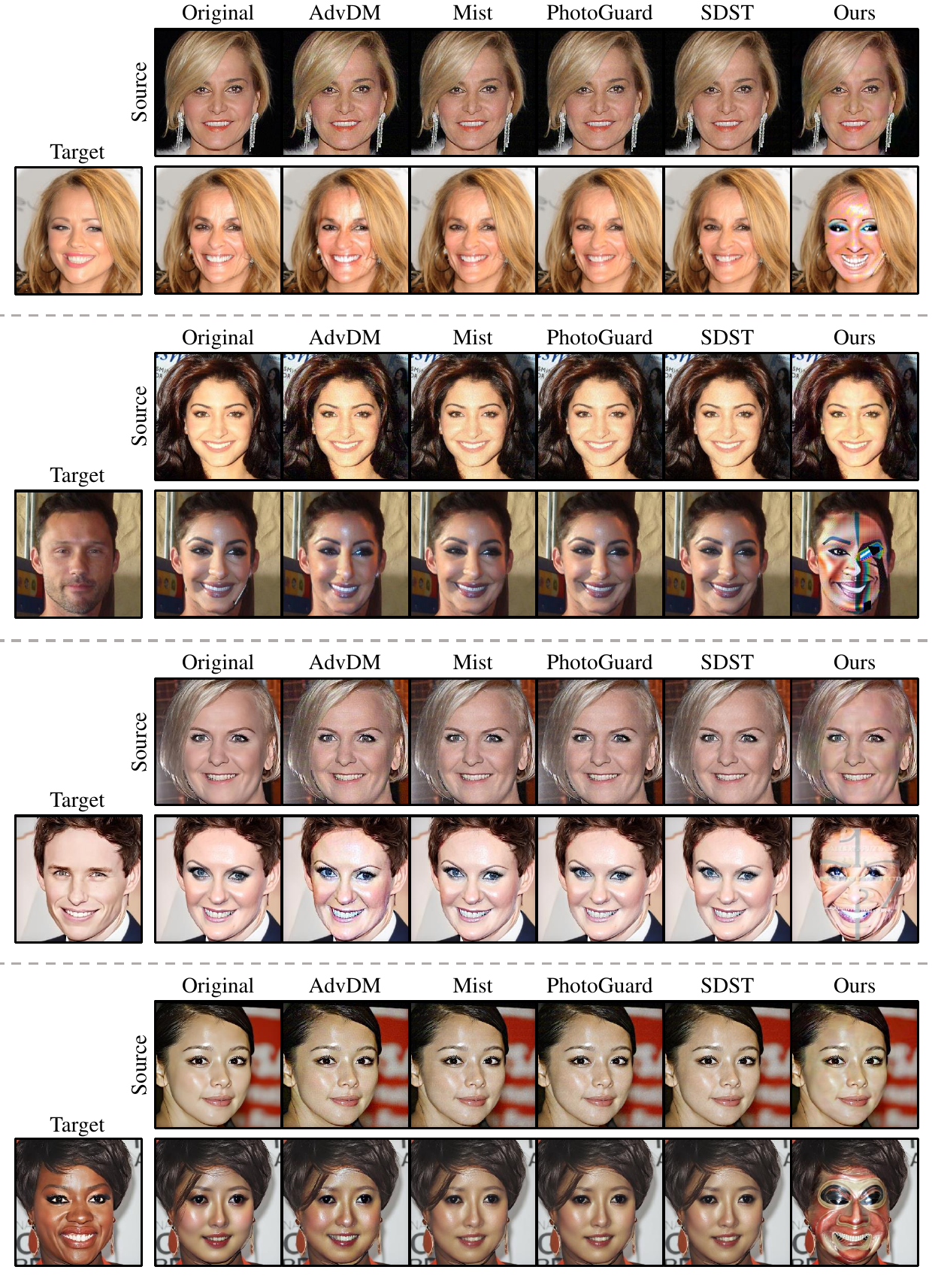}
    \caption{Qualitative comparisons for \textbf{IP-Adapter}~\cite{ye2023ip}.}
    \label{fig:supple_ipadapter}
\end{figure}

\begin{figure}[ht]
    \centering
    \includegraphics[width=0.9\columnwidth]{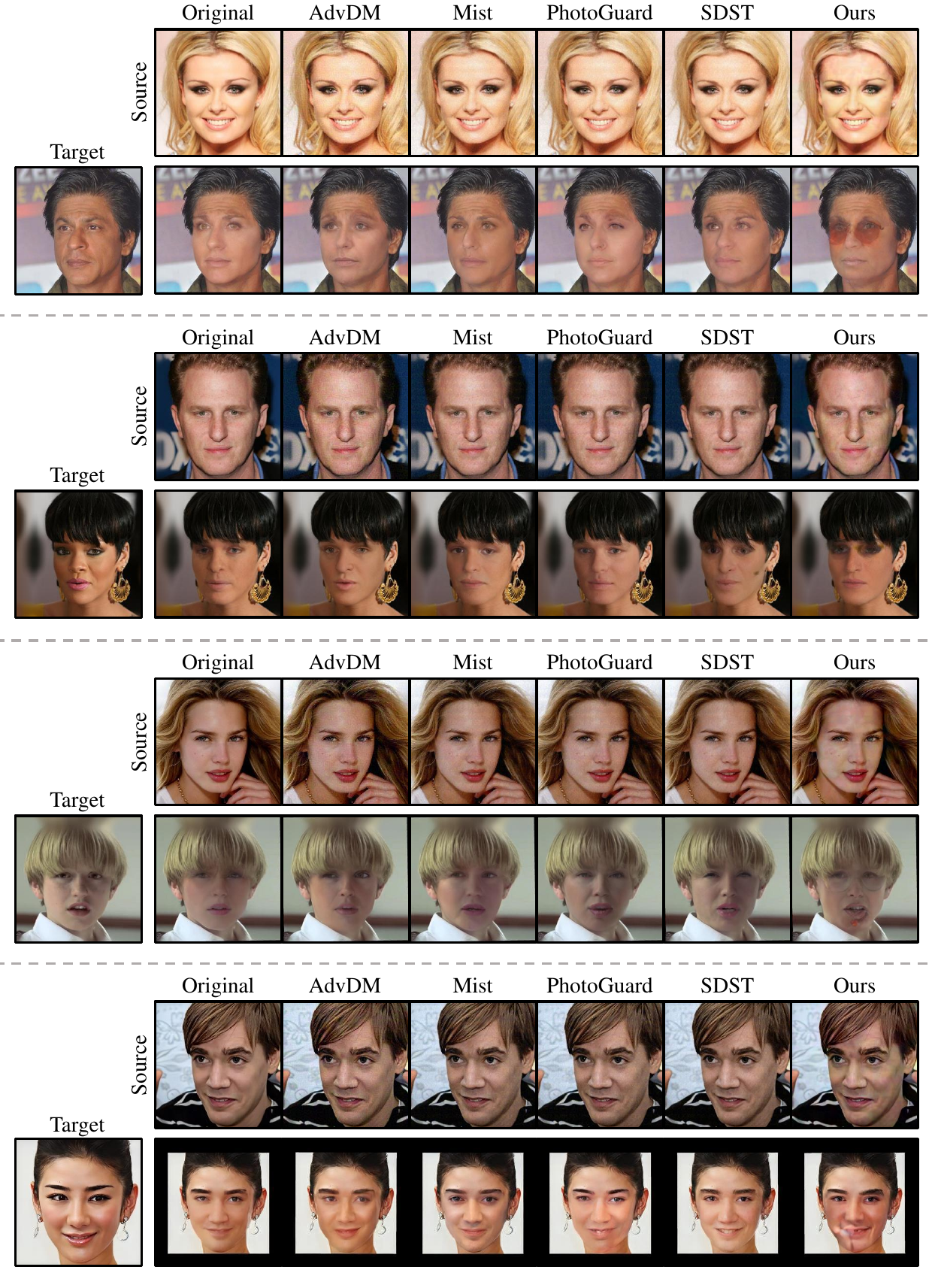}
    \caption{Qualitative comparisons for \textbf{DiffSwap}~\cite{zhao2023diffswap}.}
    \label{fig:supple_diffswap}
\end{figure}

\begin{figure}[ht]
    \centering
    \includegraphics[width=0.9\columnwidth]{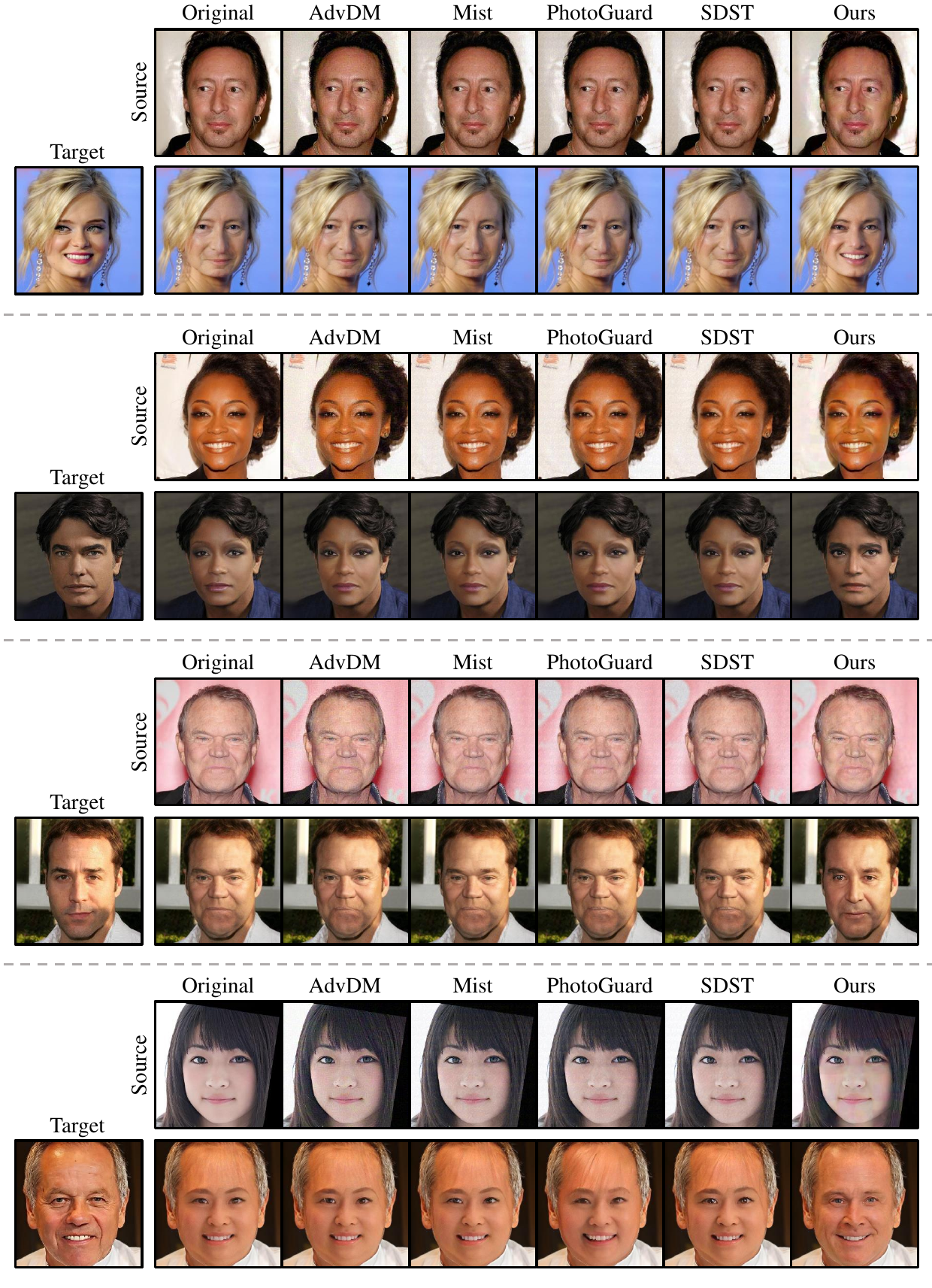}
    \caption{Qualitative comparisons for \textbf{DiffFace}~\cite{kim2212diffface}.}
    \label{fig:supple_diffface}
\end{figure}

\begin{figure}[ht]
    \centering
    \includegraphics[width=0.8\columnwidth]{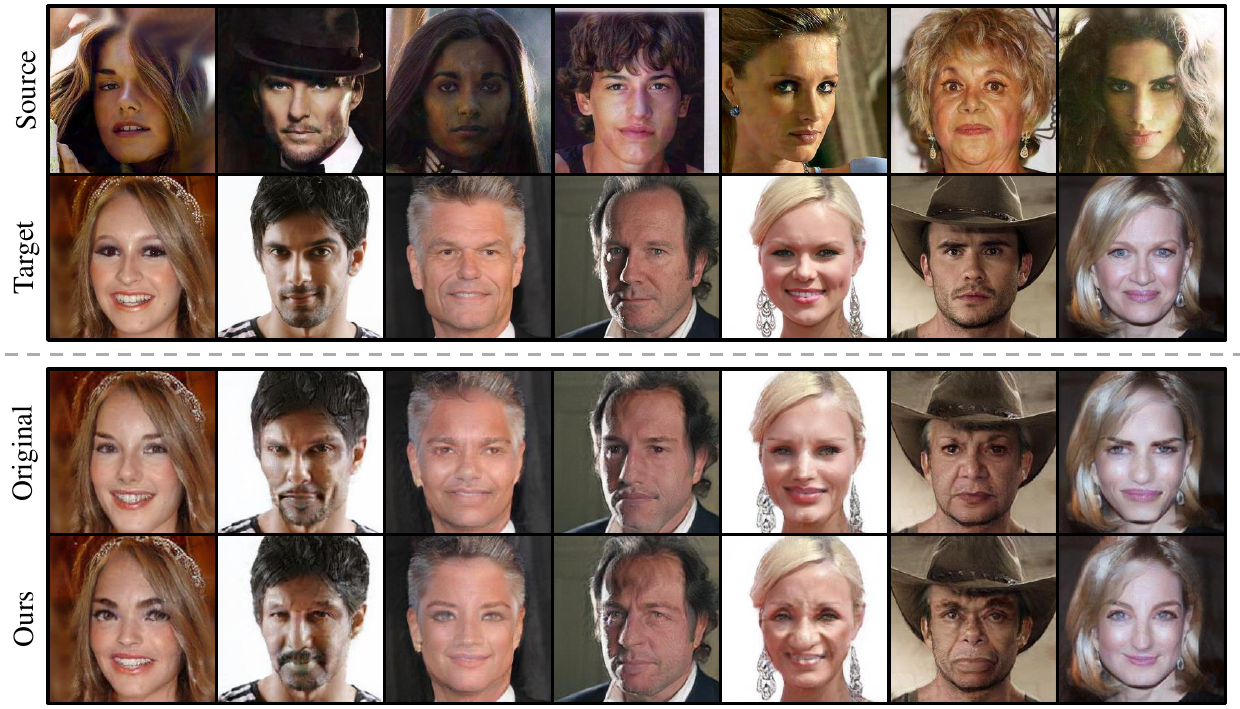}
    \caption{Qualitative results for \textbf{SimSwap}~\cite{chen2020simswap}.}
    \label{fig:supple_simswap}
\end{figure}

\begin{figure}[ht]
    \centering
    \includegraphics[width=0.8\columnwidth]{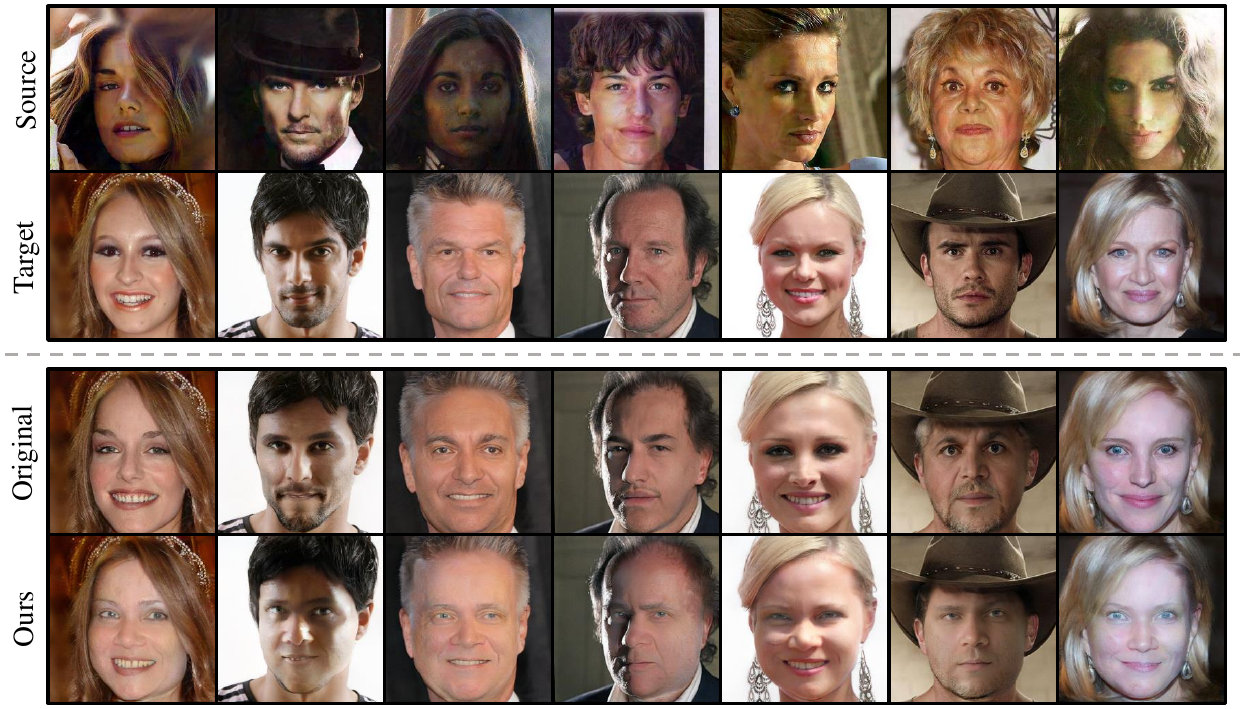}
    \caption{Qualitative results for \textbf{InfoSwap}~\cite{gao2021information}.}
    \label{fig:supple_infoswap}
\end{figure}

In this section, we present additional qualitative results of our methods. Specifically, Fig.\ref{fig:supple_fig_full_1} to Fig.\ref{fig:supple_fig_full_3} compare our approach with baseline methods~\cite{liang2023adversarial,liang2023mist,salman2023raising,xue2023toward} on various diffusion-based deepfake models~\cite{kim2212diffface,zhao2023diffswap,wang2024face,ye2023ip}, using a pair of source and target images. 
Fig.\ref{fig:supple_faceswap} compares our method with the baselines on the FaceSwap via Diffusion model~\cite{wang2024face} across different image pairs. Fig.\ref{fig:supple_ipadapter} shows the comparison within the IP-Adapter model~\cite{ye2023ip}, while Fig.\ref{fig:supple_diffswap} compares our method with the baselines on the DiffSwap model~\cite{zhao2023diffswap}. Fig.\ref{fig:supple_diffface} presents a comparison on the DiffFace model~\cite{kim2212diffface}. 
Finally, Fig.\ref{fig:supple_simswap} and Fig.\ref{fig:supple_infoswap} showcase additional experiments on two GAN-based deepfake models: SimSwap~\cite{chen2020simswap} and InfoSwap~\cite{gao2021information}, respectively.

\clearpage
\section{Additional Experiments}
\label{sup:g}
\myparagraph{Transferability experiments on variants of IP-Adapter}

\begin{figure*}[ht]
\centering
\includegraphics[width=\textwidth]{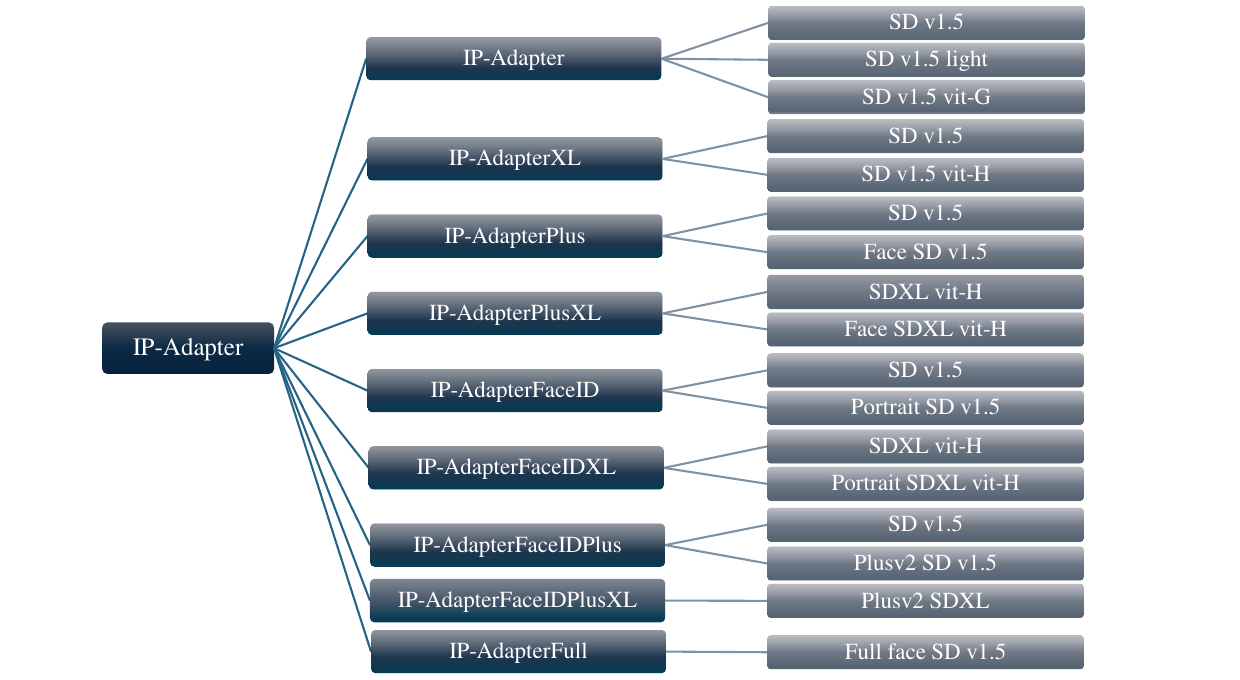}
\captionof{figure}
{\textbf{IP-Adapter model family tree}. This diagram shows the hierarchical structure of the IP-Adapter variants.}
\label{fig:ip_tree}
\end{figure*}

IP-Adapter~\cite{ye2023ip} is a lightweight adapter that enables image conditions in pre-trained text-to-image diffusion models~\cite{Rombach_2022_CVPR}.
Previous approaches~\cite{ruiz2023dreambooth, kumari2023multi} that utilized image conditions primarily relied on fine-tuning text-conditioned diffusion models. However, these methods often demanded significant computational resources and resulted in models that were challenging to reuse. 
To address these limitations, the IP-Adapter, which proposes a decoupled cross-attention mechanism, has drawn considerable attention for its practical applicability.
It is commonly used in inpainting methods with image conditions.
As shown in Fig.\ref{fig:ip_tree}, multiple versions of the IP-Adapter model have been developed with Stable diffusion v1.5~\cite{Rombach_2022_CVPR}.

A more detailed look at the various models reveals that the original model~\cite{ye2023ip} uses the CLIP image encoder~\cite{radford2021learning} to extract features from the input image. In contrast, the IP-AdapterXL improves on this by utilizing larger image encoders, such as ViT-BigG or ViT-H, which enhance both capacity and performance.
On the other hand, the IP-AdapterPlus and XL versions modify the architecture by adopting a patch embedding method inspired by Flamingo's perceiver resampler~\cite{alayrac2022flamingo}, allowing for more efficient image encoding. 
Similarly, the IP-AdapterFaceID and XL versions replace the CLIP image encoder with InsightFace, extracting FaceID embeddings from reference images. This enables the combination of additional text-based conditions with the facial features of the input image, allowing for the generation of diverse styles.
The IP-AdapterFaceIDPlus and XL versions further enhance the image encoding pipeline by incorporating multiple components. InsightFace is used for detailed facial features, the CLIP image encoder captures global facial characteristics, and the Perceiver-resampler effectively combines these features to improve the model's overall functionality.

\begin{figure}[ht]
    \centering
    \includegraphics[width=0.85\columnwidth]{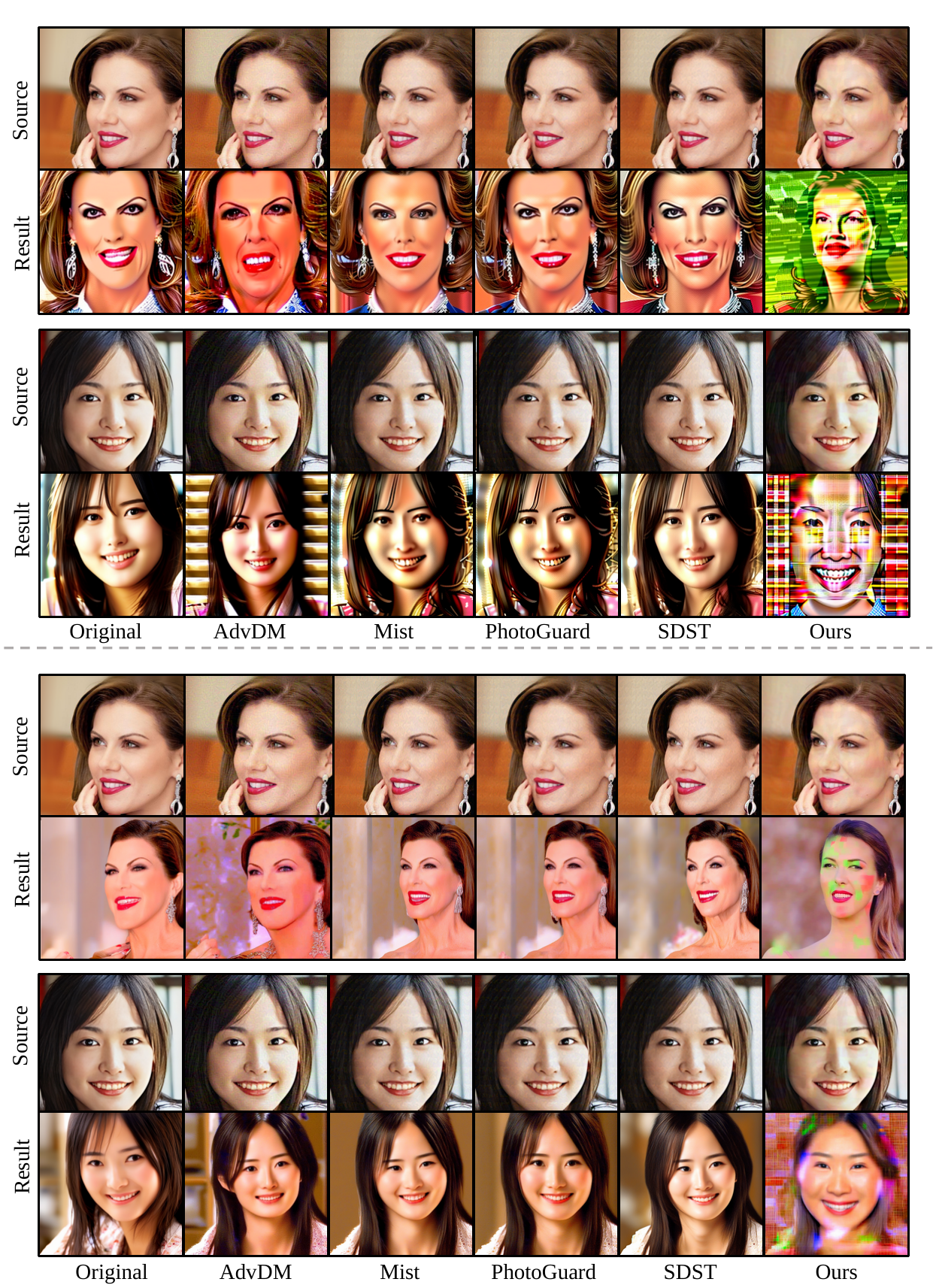}
    \caption{Qualitative comparison with baselines on the SD 1.5-based IP-Adapter \textbf{ControlNet} version (top) and SDXL-based IP-Adapter \textbf{ControlNet} version (bottom).}
    \label{fig:supple_ipadapter_controlnet}
\end{figure}

\begin{figure}[ht]
    \centering
    \includegraphics[width=0.85\columnwidth]{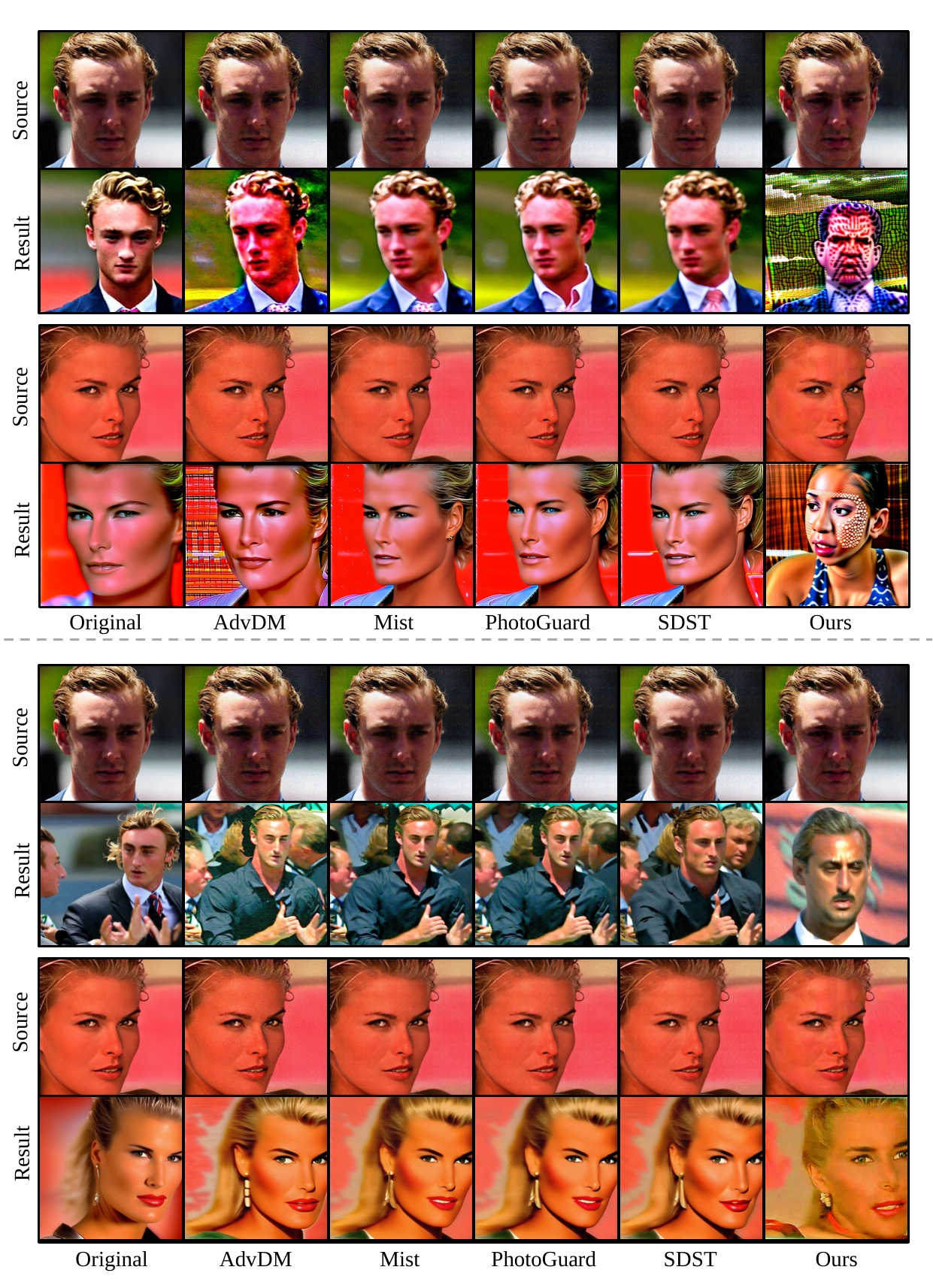}
    \caption{Qualitative comparison with baselines on the SD 1.5-based IP-Adapter \textbf{ImageVariation} version (top) and SDXL-based IP-Adapter \textbf{ImageVariation} version (bottom).}
    \label{fig:supple_ipadapter_imagevariation}
\end{figure}

\begin{figure}[ht]
    \centering
    \includegraphics[width=0.85\columnwidth]{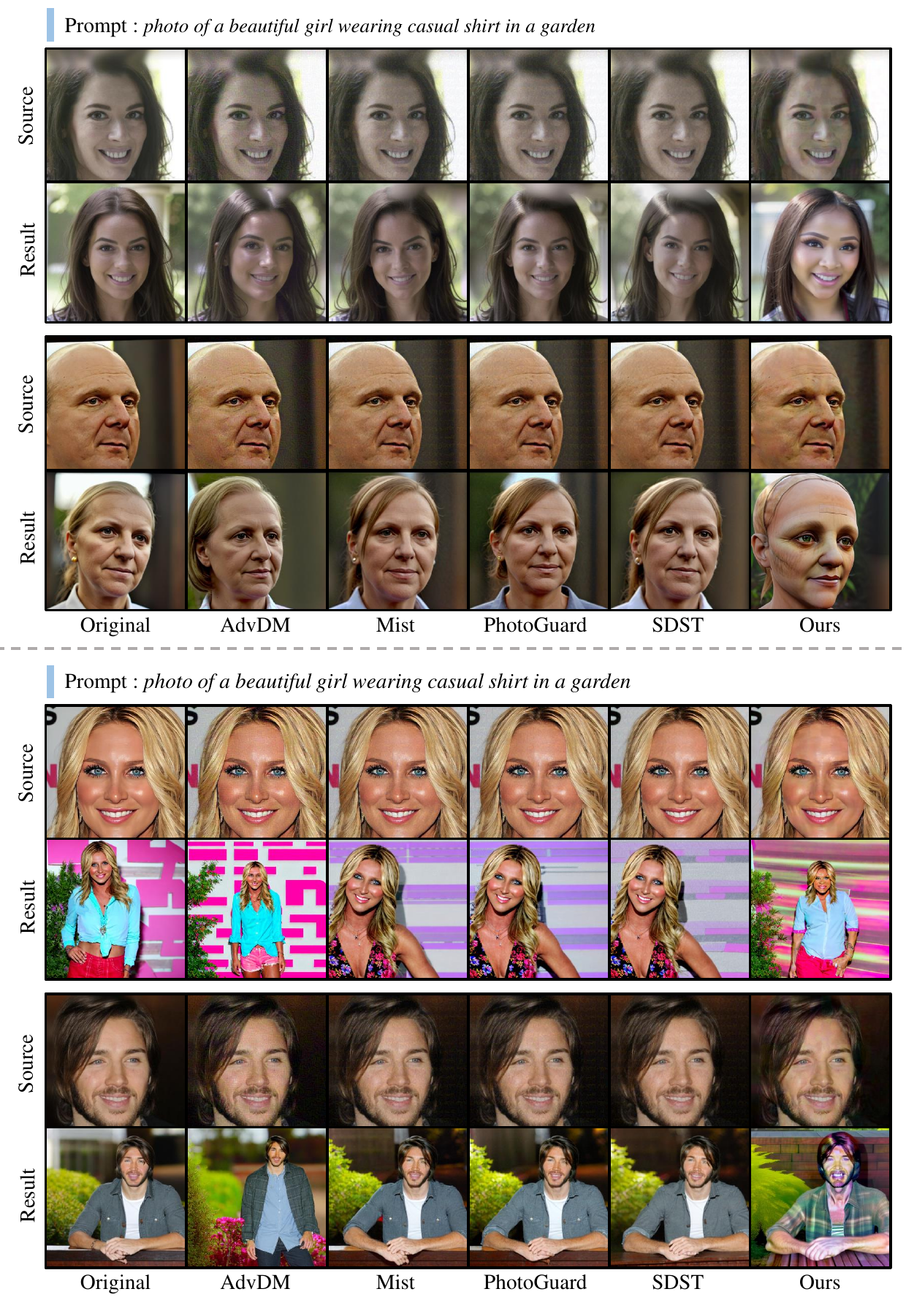}
    \caption{Qualitative comparison with baselines on the SD 1.5-based IP-Adapter \textbf{Multi-modal prompts} version (top) and SDXL-based IP-Adapter \textbf{Multi-modal prompts} version (bottom).}
    \label{fig:supple_ipadapter_multimodalprompts}
\end{figure}

\begin{figure}[ht]
    \centering
    \includegraphics[width=0.85\columnwidth]{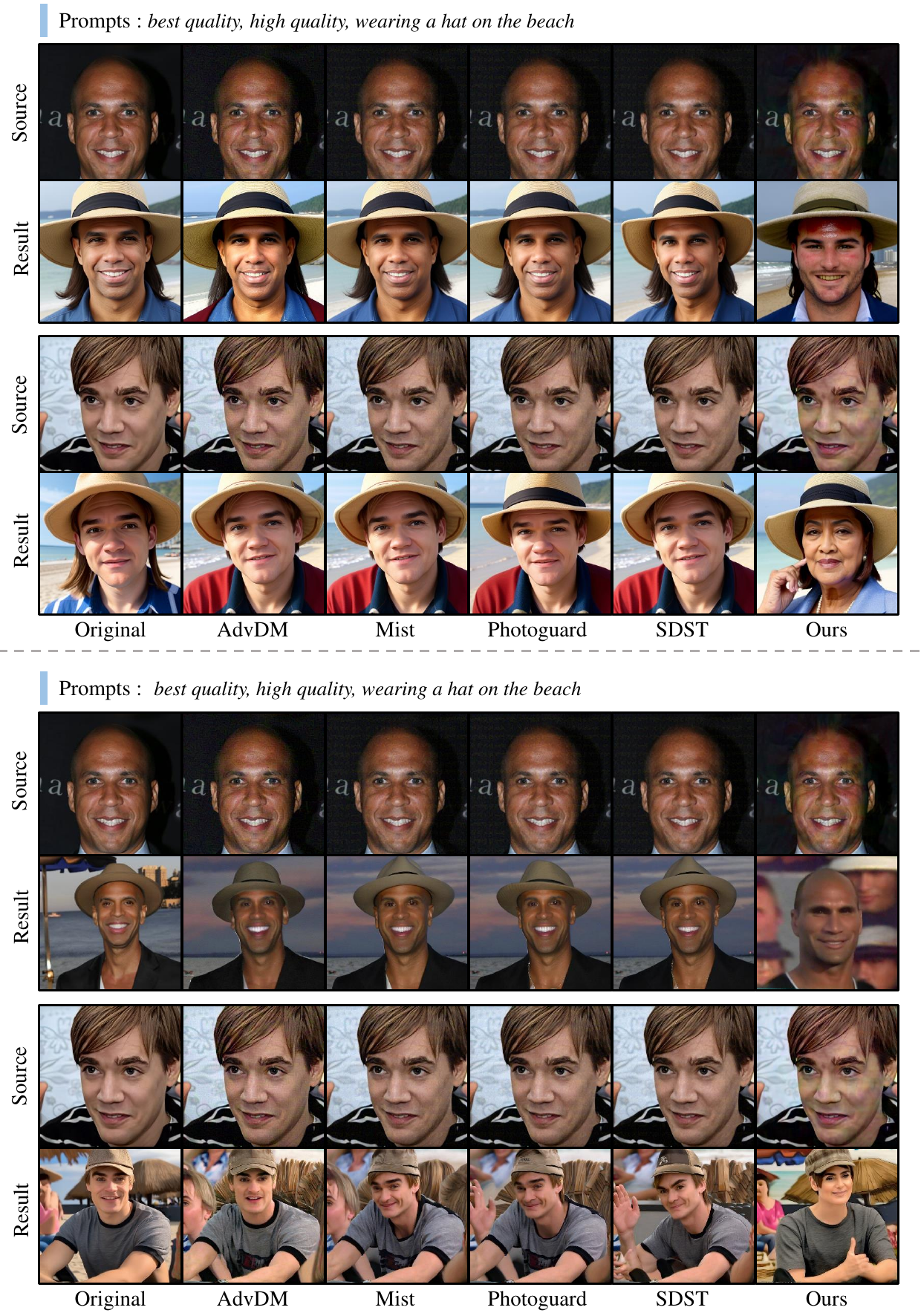}
    \caption{Qualitative comparison with baselines on the SD 1.5-based IP-Adapter \textbf{Plus} version (top) and the SDXL-based IP-Adapter \textbf{Plus Face} version (bottom).}
    \label{fig:supple_ipadapter_plus}
\end{figure}

\myparagraph{Qualitative results.}
We evaluate the transferability across different IP-Adapter versions and present comparisons with baseline methods.
Specifically, we conducted experiments on eight of these models, with results and model descriptions provided in Fig.\ref{fig:supple_ipadapter_controlnet} to Fig.\ref{fig:supple_ipadapter_plus}. 
These results demonstrate the versatility of \textit{FaceShield}, showing that it is applicable across various sub-models of the IP-Adapter~\cite{ye2023ip}.

\end{document}